\RequirePackage{fix-cm}
\documentclass[smallextended]{svjour3}       
\smartqed  

\usepackage{graphicx}
\usepackage{amssymb}
\usepackage{geometry}
\usepackage{subfig}
\usepackage{caption}
\usepackage{algorithm}
\usepackage{algpseudocode}
\usepackage{amsmath}
\usepackage[numbers]{natbib}
\usepackage{multirow}
\usepackage{makecell}
\usepackage{setspace}
\usepackage[colorlinks,
            linkcolor=blue,
            anchorcolor=blue,
            citecolor=blue,
            urlcolor=blue,
            ]{hyperref}
\usepackage[misc]{ifsym}
\newcommand{\tabincell}[2]{\begin{tabular}{@{}#1@{}}#2\end{tabular}} %
\begin{document}
\def\*#1{\mathbf{#1}}
\title{Adaptive Multi-layer Contrastive Graph Neural Networks
}


\author{{Shuhao Shi}    \textsuperscript{1} \and
        {Pengfei Xie}   \textsuperscript{1} \and
        {Xu Luo}        \textsuperscript{1} \and
        {Kai Qiao}       \textsuperscript{1}\and
        {Linyuan Wang}   \textsuperscript{1}\and
        {Jian Chen}      \textsuperscript{1}\and
        {Bin Yan}        \textsuperscript{1,~\Letter}
}


\institute{
    \begin{itemize}
      \item[] {Shuhao Shi} \\
            \email{ssh$\_$smile@163.com}
      \\
       \item[]{Pengfei Xie} \\
           \email{Luokysicy@outlook.com}
      \\
           \item[]{Xu Luo} \\
           \email{bird831037@163.com}
      \\
           \item[]{Kai Qiao} \\
           \email{qiaokai1992@gmail.com}
      \\
           \item[]{Linyuan Wang} \\
           \email{wanglinyuanwly@163.com}
      \\
           \item[]{Jian Chen} \\
           \email{kronhugo@163.com}
      \\
           \item[\textsuperscript{\Letter}]{Bin Yan} \\
           \email{ybspace@hotmail.com}
      \at
      \item[\textsuperscript{1}] Henan Key Laboratory of Imaging and Intelligence Processing, PLA strategy support force information engineering university, Zhengzhou, China
    \end{itemize}
}

\date{Received: date / Accepted: date}

\maketitle

\begin{abstract}
We present Adaptive Multi-layer Contrastive Graph Neural Networks (AMC-GNN), a self-supervised learning framework for Graph Neural Network, which learns feature representations of sample data without data labels. AMC-GNN generates two graph views by data augmentation and compares different layers' output embeddings of Graph Neural Network encoders to obtain feature representations, which could be used for downstream tasks. AMC-GNN could learn the importance weights of embeddings in different layers adaptively through the attention mechanism, and an auxiliary encoder is introduced to train graph contrastive encoders better. The accuracy is improved by maximizing the representation's consistency of positive pairs in the early layers and the final embedding space. Our experiments show that the results can be consistently improved by using the AMC-GNN framework, across four established graph benchmarks: Cora, Citeseer, Pubmed, DBLP citation network datasets, as well as four newly proposed datasets: Co-author-CS, Co-author-Physics, Amazon-Computers, Amazon-Photo.

\end{abstract}

\section{Introduction}
\label{intro}
Graph Neural Networks (GNNs) are effective methods for analyzing graph data, and various downstream graph learning tasks such as node classification, similarity search, graph classification, and link prediction have benefited from its recent developments \cite{article15,article21,article22}. However, most of the existing GNNs frameworks are supervised learning methods that have many drawbacks \cite{article15,article21,article22,article34}. GNNs obtained by supervised learning tend to learn task-specific knowledge, and the learned feature representations are difficult to transfer to other tasks \cite{article4}. Besides, supervised learning requires labeled data as input, and the over-reliance on labeled information will lead to poor robustness. Moreover, obtaining labeled information for large amounts of training data is labor-intensive, especially in the presence of large-scale networks.
\par In recent years, there has been a tremendous development in graph contrastive learning. Contrastive learning adopts data augmentation to obtain semantically identical features and then maximize feature consistency across augmented views to learn the representation. Researchers initially applied the contrast learning framework in computer vision to overcome the drawbacks, , such as the need for data annotation and the poor transferability of the learned features \cite{article1,article4,article24,article25,article27,article31}.
\par Graph data augmentation is not simple to define in contrastive learning methods, in contrast to diverse data transformation techniques for images and text. Graph data augmentation is more complex because of the non-Euclidean nature of graphs \cite{article32}, and the existing research mainly focuses on improvements to graph data augmentation. Inspired by contrastive learning in images such as MOCO \cite{article24}, SimCLR \cite{article4}, and DIM \cite{article27}, \itshape etc\upshape., many methods in graph contrastive learning have been proposed.
Many studies involve data augmentation by changing the edges and nodes of the graph. GCC \cite{article38} performing random walks in the network to construct subgraphs about nodes for contrastive learning. CGNN \cite{article11} learns consistent representations of nodes through different sampled neighbors. GraphCL \cite{article10} designs four types of graph data augmentation: Node dropping, Edge perturbation, Attribute masking, and Subgraph, which are combined with various prior knowledge to select data expansion methods in practical application. GCA \cite{article2} designs augmentation schemes based on node centrality measures to highlight important connective structures on the topology level.
\par Some studies perform data augmentation through constructing local-global pairs and negative-sampled counterparts. DGI \cite{article32} and InfoGraph \cite{article37} migrates DIM \cite{article27} to graphs and propose unsupervised learning objectives based on Mutual Information (MI). GMI \cite{article13} focuses on arriving at graphical mutual information maximization in a node-level by directly maximizing MI between inputs and outputs of the encoder. \cite{article36,article18} further extends MI maximization to heterogeneous graphs. The unsupervised learning model of graph structure is trained by maximizing the MI between the graph’s local features and global features. GRACE \cite{article12} simplifies DGI \cite{article32} by obtaining graph node representations by maximizing node embeddings’ consistency between two graph views generated through structure and feature perturbations. GRACE even surpasses its supervised counterparts on transductive tasks.
\par  Different from previous graph contrastive learning models, we improve the framework structures of graph contrastive learning rather than graph data augmentation methods. The idea behind our strategy is to align semantically identical graph view data in different latent spaces to obtain a more general and differentiated representation. Then train a superior graph contrastive learning model, where the importance weights of the embeddings in different latent spaces are learned adaptively by the attention mechanism. Inspired by hierarchy semantic alignment strategy strategy on Convolutional Neural Network (CNN) \cite{article17} and OhmNet \cite{article20}, a hierarchy‐aware unsupervised node feature learning approach for multi‐layer networks, we propose a contrasting framework for unsupervised graph representation learning, called Adaptive Multi-layer Contrastive Graph Neural Network (AMC-GNN), which introduces adaptive multi-layer contrastive loss into graph contrastive learning models.

Specifically, semantically identical graph data is first obtained by graph data augmentation, and then the consistency between the node embeddings of the two graph views is maximized by minimizing the adaptive multi-layer contrastive loss. By optimizing the embedding consistency of the middle layer and the final embedding space, the representation consistency of the embedding generated by the GNN encoder is improved. In addition, we also introduce an auxiliary training model to improve the performance of the model further.
\par Our contributions are as follows:
\par 1. We propose the adaptive multi-layer graph contrastive learning framework that can be generalized to existing GNN models. Extensive experiments have been conducted to demonstrate that AMC-GNN can provide comparable or better performance than supervised models in graph data tasks.
\par 2. We experimentally proved that AMC-GNN has stronger robustness than other unsupervised models under slight perturbation. And we also verified that AMC-GNN has better performance at different feature removal rates.
\par 3. We have explored the introduction of adaptive multi-layer contrastive loss and auxiliary models to improve performance through ablation experiments, and verified that adaptive multi-layer contrastive loss is the key to improved performance.

\section{Related works}
\label{sec:1}
\subsection{Contrastive Learning}
\label{sec:2}
Contrastive learning is a self-supervised learning method whose main idea is to train the feature encoder by making the positive samples as close as possible and the negative samples as far away as possible in the feature space. The effect of only using a single data augmentation method on learning representation is general \cite{article4}, so well-constructed embeddings is essential for learning good representations. In the field of computer vision, many studies have been conducted on data augmentation for contrastive learning \cite{article4,article24,article25,article27}. Generally, the same image is rotated, cropped, divided into subgraphs, and other transformations that do not change the image’s semantics to construct its positive sample pairs \cite{article1,article4,article31}. DIM \cite{article27}, AMDIM \cite{article28} uses the principle of maximizing MI to maximize the MI between local features and global features of the same image. SimCLR \cite{article4} proposed Projection Head, and SimCLR v2 \cite{article31} verified that a deeper Projection Head could improve the quality of feature representation, and similar structures were introduced in subsequent studies \cite{article24,article25}. For data such as text and audio, samples within a certain window are usually considered as positive pairs \cite{article29,article30}, considering their temporal order. In graph contrastive learning, graph data is not as informative as the geometric and structured information that images have, and most research has focused on exploring the augmentation of structured graph data \cite{article10}.

\subsection{Graph Representation Learning}
\label{sec:3}
Traditional unsupervised graph representation learning node2vec \cite{article14}, Deepwalk \cite{article19}, VGAE \cite{article33} focuses on local contrast, forcing neighboring nodes to have similar embeddings. These approaches over-emphasize the structural information encoded in graph proximity and have the disadvantage of being difficult to handle large-scale datasets \cite{article14,article19}. Recent work on graphs employs graph convolutional network (GCN) encoders that better than conventional methods.
In the graph field, most of the research has studied supervised models \cite{article15,article21,article22,article34}. The rise of contrastive learning has motivated great interest in studying unsupervised learning in GNNs \cite{article9,article32,article10,article11,article12,article38}. Many methods study data augmentation by changing the edges and nodes of the graph \cite{article38,article10,article11,article2}.
A series of contrastive learning methods seeking to maximize the Mutual Information (MI) between the input and its representations have been proposed. Inspired by DIM \cite{article27} and AMDIM \cite{article28}, DGI \cite{article32} propose contrasting learning between local and global representations on graphs to capture structural information better. HDGI \cite{article36}, DMGI \cite{article18} further combine the MI with the meta-paths in the heterogeneous network, learns the weights of different meta-paths, and fuses them to obtain the final graph node representation. GRACE \cite{article12} simplifies DGI \cite{article32} by obtaining graph node representations by maximizing node embeddings’ consistency between two graph views generated through structure and feature perturbations. GRACE highlight the importance of appropriately choosing negative samples, which is often neglected in previous InfoMax-based methods.
\par Previous work focused on graph data augmentation methods. However, the information contained in the latent spaces of the middle layers is ignored. Therefore, we propose a graph contrastive learning framework that enables embeddings to be closer in multiple feature spaces.

\section{Methodology}
\label{sec:4}
\subsection{Preliminaries}
\label{sec:5}
In unsupervised graph representation learning, $\mathcal{G}=(\mathcal{V},\mathcal{E})$ denotes the undirected graph, $\mathcal{V}=\left\{ {{v}_{i}}\left| 1\le i\le N \right. \right\}$ represents the node set , and $\mathcal{E}=\left\{ {{e}_{ij}}\left| 1\le i,j\le N \right. \right\}$ represents the edge set. We define the feature matrix and adjacency matrix as $\mathbf{X}\in {{R}^{N\times F}}$ and $\mathbf{A}\in {{R}^{N\times N}}$, respectively, where ${\*{{A}}_{i,j}}=1$, if $({{v}_{i}},{{v}_{j}})\in \mathcal{E}$, while ${\*{{A}}_{i,j}}=0$ means not. Our purpose is to learn a graph encoder that $f:{{R}^{N\times F}}\times {{R}^{N\times N}}\to {{R}^{N\times {F}'}}$ such that $f(\*{X},\*{A})=\*{H}=\left\{ {\*{h}_{1}},{\mathbf{h}_{2}},...,{\*{h}_{N}} \right\}$, where ${\*{h}_{i}}$ represents the embedding of node ${{v}_{i}}$, which can be used for downstream tasks such as node classification and community detection.

\subsection{Adaptive Multi-level Contrastive Graph Neural Networks}
\label{sec:6}
Graph contrastive learning framework generally composed of three components: Data Augmentation, Encoder, and Loss. The proposed AMC-GNN introduced the auxiliary training model and the adaptive multi-layer contrastive loss in Encoder and Loss compared to the previous models. Next, we will illustrate each component of AMC-GNN and the learning process in detail.

\subsubsection{Data Augmentation for Graphs}
\label{sec:7}
The generation of different positive and negative sample pairs is an essential part of contrastive learning, because efficient positive and negative sample pairs can provide the most informative representation for downstream classification tasks.

\paragraph{Random Sampling Neighbours(RN)}~{}
\newline
For any node ${{v}_{i}}$, use $\mathcal{N}(i)$ to denote its original neighborhood specified by GNN architectures. When aggregating neighborhood nodes, randomly select neighboring nodes in $\mathcal{N}(i)$ for aggregation. To obtain different neighborhoods, we randomly dropped edges with proportion $\rho $ based on the original neighborhood to obtain new neighborhoods, denoted by $D\left( \mathcal{N}(i),\rho  \right)$. In graph contrastive learning, the best performance is achieved by comparing first-order neighbor coding and graph diffusion \cite{article5}. Therefore, we obtained two subneighborhoods $\mathcal{N}{{(i)}_{1}}$ and $\mathcal{N}{{(i)}_{2}}$ by two independent random sampling:

\begin{equation}
\label{equ:1}
\mathcal{N}{{(i)}_{1}},\mathcal{N}{{(i)}_{2}}\sim D\left( \mathcal{N}(i),\rho  \right).
\end{equation}

To improve the computational efficiency, we used a simplified approach while performing random discarding of the entire graph connection. The original graph is randomly dropped twice independently to obtain two different graphs ${{\mathcal{G}}_{1}}$ and ${{\mathcal{G}}_{2}}$. The neighborhoods of any node ${{v}_{0}}$ in the new graphs ${{\mathcal{G}}_{1}}$ and ${{\mathcal{G}}_{2}}$, are used as $\mathcal{N}{{(i)}_{1}}$ and $\mathcal{N}{{(i)}_{2}}$.

\paragraph{Attribute Masking(AM)}~{}
\newline
We randomly mask some of the dimensions in the node features with zeros, and the proportion of the masked dimensions is $p$. Formally, we first construct a random mask vector $\*{m}\in {{\left\{ 0,1 \right\}}^{F}}$, with each dimensional component of $\*m$ independently drawn from a Bernoulli distribution with probability $1-p$, and ${{m}_{i}}$ is the $i$th component of the vector, where ${{m}_{i}}\sim \mathcal{B}\left( 1-p \right)$, $1\le i\le N$. Use $\*{\tilde{X}}$ to denote the feature matrix $\*X$ processed by the AM,

\begin{equation}
\label{equ:2}
\*X= [{\*{{x}}_{1}},{\*{{x}}_{2}},...,{\*{{x}}_{N}}] , \*{\tilde{X}}= [{{\*{\tilde{x}}}_{1}},{{\*{\tilde{x}}}_{2}},...,{{\*{\tilde{x}}}_{N}}],
\end{equation}

\noindent where ${{\*{\tilde{x}}}_{i}}={\*{x}_{i}}\circ \*m$, $\left( \circ  \right)$ denotes the inner product.
\par Our framework jointly utilizes both RN and AM methods to obtain semantically identical graph data for subsequent graph contrastive learning. The generation of two semantically identical graph data ${{\mathcal{G}}_{1}}$ and ${{\mathcal{G}}_{2}}$ is affected by the hyperparameters ${{p}_{R,1}}$, ${{p}_{A,1}}$, ${{q}_{R,1}}$, ${{q}_{A,1}}$ and ${{p}_{R,2}}$, ${{p}_{A,2}}$, ${{q}_{R,2}}$, ${{q}_{A,2}}$. ${{p}_{*}}$, ${{q}_{*}}$ represent the hyperparameters for target and auxiliary encoder data augmentation, respectively, and the subscripts R and A denote RN and AM.

\subsubsection{GNN-based encoder}
\label{sec:8}
In our model, two GNNs in series are used as encoders, where the first model is the target encoder and the second model is the auxiliary training encoder. The idea behind this is to increase the number of layers of the encoder to generate more embeddings of different layers as input to the adaptive multi-layer contractive loss, adding more constraints in the process of target encoder optimization. ${{l}_{1}}$ and ${{l}_{2}}$ indicate the number of layers of the target and auxiliary models, respectively, and the general number of layers of the encoders are two. The first model’s output embeddings are used as the final outputs and the feature vectors for the second model’s input. Let ${{f}_{1}}(\cdot )$ and ${{f}_{2}}(\cdot )$ be target model and auxiliary training model, then feature representation in different layers could get through Eq. \ref{equ:3}.

\begin{equation}
\label{equ:3}
{\*{{h}}^{n}}=f_{1}^{n}(\*{\tilde{X}},\*{\tilde{A}}), {\*{{h}}^{m}}=f_{2}^{m-{{l}_{1}}}({\*{{h}}^{{{l}_{1}}}},{\*{\tilde{A}}}'),
\end{equation}

\noindent where $f_{*}^{k}(\cdot )$ means passing through the $k$-layer GNN, $1\le n\le {{l}_{1}}, {{l}_{1}}<m\le {{l}_{1}}+{{l}_{2}}$, $\*{\tilde{A}}$ and $\*{{\tilde{A}}}'$, respectively, formed from $\mathbf{A}$ after two independent RN.

\begin{figure}
\centering
  \includegraphics[width=0.50\textwidth]{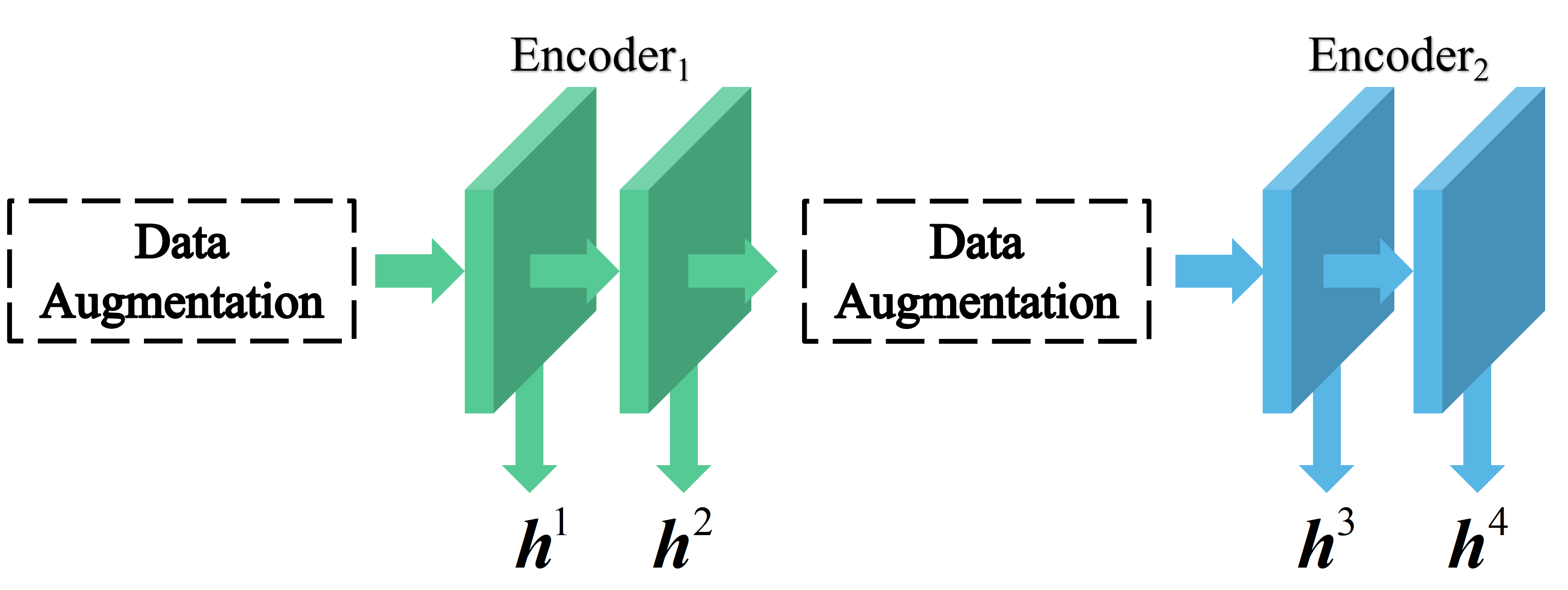}
\caption{The GNN-based encoder.}
\label{fig:1}       
\end{figure}

\par For the $i$th node, where $1\le i\le N$, the node after one graph transformation is ${{u}_{i}}$ and after another graph transformation is ${{v}_{i}}$. Nodes ${{u}_{i}}$ and ${{v}_{i}}$ form embedded vectors $\*{h}_{{{u}_{i}}}^{k}$ and $\*{h}_{{{v}_{i}}}^{k}$ respectively after passing through $k$-layers of GNN. Then the neural network Projection Head is used to map the embedding to the contrast space to obtain the vectors $\*{z}_{{{u}_{i}}}^{k}$ and $\*{z}_{{{v}_{i}}}^{k}$, where $1\le k\le {{l}_{1}}+{{l}_{2}}$. The Projection Head is composed of a 2-layer multi-layer perceptron (MLP) to enhance the expression power of the critic \cite{article17,article26} to avoid the loss function that computes similarity from dropping some important features during training \cite{article4}. The process can be represented as:$\*z={\*{W}_{2}}\left( \sigma \left( {\*{W}_{1}}*\*{h} \right) \right)$, where $\sigma \left( \cdot  \right)$ is the activation function and $\*W$ is linear layer.

\begin{figure}
\centering
  \includegraphics[width=0.65\textwidth]{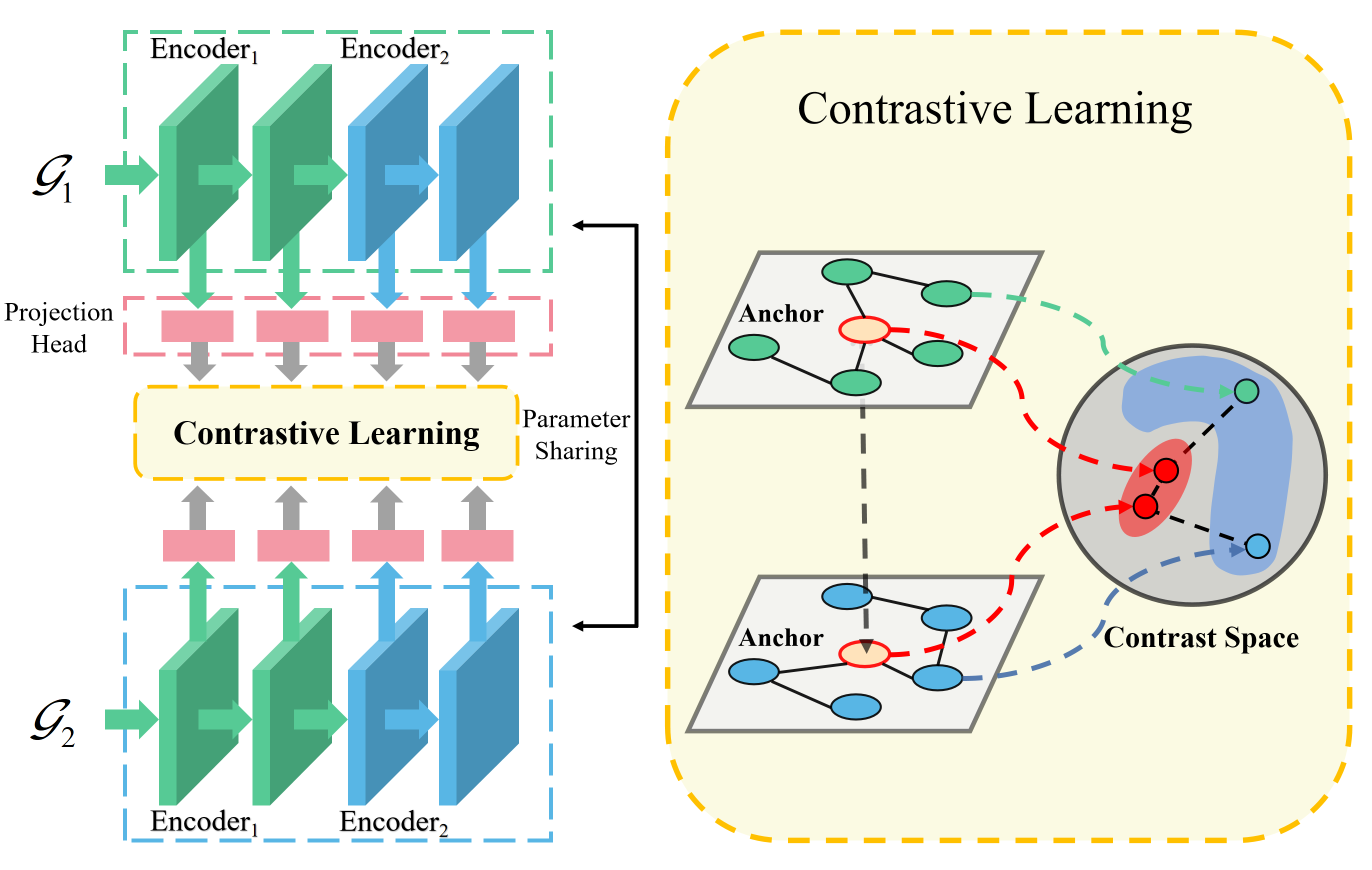}
\caption{The illustrative schematic diagram of our proposed framework AMC-GNN. The model consists of three components: 1. GNN-based encoder: two GCN encoders are used to generate the feature representation vectors; 2. Projection Heads: four different projection heads are used to project the resulting embedding vectors into the loss space; 3. Contrastive loss: calculates the sum of positive and negative sample contrastive losses for the middle layer as well as the final embedding space.}
\label{fig:2}       
\end{figure}

\subsubsection{Adaptive Multi-layer Contrastive Loss}
\label{sec:9}
Different from previous graph contrastive learning algorithms, we extend the contrastive loss to learn a more distinguishing feature representation in different layers. Specifically, the previous contrastive GNN model uses only the embedding layer vectors generated in the last layer after training the encoder, while the optimization of the middle hidden layers is only performed by backpropagating the gradient to the earlier layers. In this case, we extend the contrastive loss proposed in \cite{article12} by adding each layer’s embeddings to the contrastive loss. Combined with the attention mechanism, information from different layers can be adequately fused.
\par We learn node embeddings by maximizing node-level agreement between embeddings in different layers. By minimizing the node embedding’s loss, the coded embedding of each node after two different transformations is made consistent while being far away from the other nodes coded embedding in the feature space. $\*{u}_{i}^{k}$ and $\*{v}_{i}^{k}$ are the embeddings at the output of $k$th layer of the GNN encoder and their loss functions are shown in Eq. \ref{equ:5}.

\begin{small}
\begin{equation}
\label{equ:5}
\mathcal{L}\left( \*{z}_{{{u}_{i}}}^{k},\*{z}_{{{v}_{i}}}^{k} \right)=\log \frac{\exp \left( s(\*{z}_{{{u}_{i}}}^{k},\*{z}_{{{v}_{i}}}^{k})/\tau  \right)}{\exp \left( s(\*{z}_{{{u}_{i}}}^{k},\*{z}_{{{v}_{i}}}^{k})/\tau  \right)+\sum\limits_{j=1}^{N}{{{\mathbb{I}}_{[i\ne j]}}\exp \left( s(\*{z}_{{{u}_{i}}}^{k},\*{z}_{{{v}_{j}}}^{k})/\tau  \right)}+\sum\limits_{j=1}^{N}{{{\mathbb{I}}_{[i\ne j]}}\exp \left( s(\*{z}_{{{u}_{i}}}^{k},\*{z}_{{{u}_{j}}}^{k})/\tau  \right)}},
\end{equation}
\end{small}

\noindent where $s(\*{x},\*{y})={\*{(x)}}^{T}\*{y}$ calculates the similarity of two vectors, and $\tau $ is an adjustable factor. ${{\mathbb{I}}_{[i\ne j]}}$ represents the indicator function, which is 1 when $i$ and $j$ are not equal and 0 otherwise. The numerator of the loss function is the positive pair, which encourages similar vectors to be close together, and the denominator is the positive pair and negative pairs, which pushes all other vectors apart from the positive pair. The loss function of $k$th-layer is calculated by Eq. \ref{equ:8-1}.

\begin{equation}
\label{equ:8-1}
\mathcal{L}^{k}=\frac{1}{2N}\sum\limits_{i=1}^{N}{\left[ \mathcal{L}\left( \*{z}_{{{u}_{i}}}^{k},\*{z}_{{{v}_{i}}}^{k} \right)+\mathcal{L}\left( \*{z}_{{{v}_{i}}}^{k},\*{z}_{{{u}_{i}}}^{k} \right) \right]}.
\end{equation}

Adding the losses of different layers directly does not necessarily adapt to different datasets. We learn the weights of the embedding losses of different layers adaptively by attention mechanism. We firstly transform the embedding through a nonlinear transformation, and then use one shared attention vector ${{\mathbf{q}}^{k}}\in {{{R}}^{N\times 1}}$ to get the attention value $\omega _{{{u}_{i}}}^{k}$ as follows:

\begin{equation}
\label{equ:6}
\omega _{{{u}_{i}}}^{k}={{({{\mathbf{q}}^{k}})}^{T}}\cdot \tanh ({{\mathbf{W}}^{k}}\cdot {{(\mathbf{z}_{{{u}_{i}}}^{k})}^{T}}+\mathbf{b}),
\end{equation}

\noindent Here $\mathbf{W}^{k}\in {{R}^{N\times F}}$ is the weight matrix and $\mathbf{b}\in {{R}^{N\times 1}}$ is the bias vector. We then normalize the attention values $\omega _{{{u}_{i}}}^{k}$ with softmax function to get the final weight:

\begin{equation}
\label{equ:7}
\alpha _{{{u}_{i}}}^{k}=\text{soft}\max (\omega _{{{u}_{i}}}^{k})=\frac{\exp (\omega _{{{u}_{i}}}^{k})}{\sum\limits_{k=1}^{M}{\exp (\omega _{{{u}_{i}}}^{k})}},
\end{equation}

\noindent Larger $\alpha _{{{u}_{i}}}^{k}$ implies the corresponding embedding is more important. Finally, the loss of all positive sample pairs is calculated to obtain the overall loss as in Eq. \ref{equ:8-2}.

\begin{equation}
\label{equ:8-2}
\mathcal{{L}}_{total}={\sum\limits_{k=1}^{M}{\alpha _{{{u}_{i}}}^{k}\cdot \mathcal{{L}}^{k}}}.
\end{equation}

By minimizing $\mathcal{L}_{total}$, the effect of maximizing the lower bound on the MI between positive sample pairs can be achieved \cite{article1}. The previous graph contrastive learning model uses only the embeddings generated in the last layer, while the optimization of the middle hidden layers is performed only by back-propagating the gradients to the earlier layers. The optimization of the middle layers is difficult to converge to the optimization objective due to the absence of labels. We use the adaptive multi-layer contrastive loss to learn the feature representations in different layers, ensuring that the middle layers are also well optimized, thereby enhancing the model’s performance. Though our method capitalizes on multiple layers, they are all part of the same model, therefore, incur no additional computational overhead in reasoning.

\begin{algorithm}[htbp]\small
\begin{spacing}{1.3}
  \caption{Main steps of the AMC-GNN algorithm.}
  \label{alg:Framwork}
  \begin{algorithmic}[1]
    \For{$epoch\leftarrow 1,2,...,N$};
    \label{code:fram}
      \State Generate two graph view ${{\mathcal{G}}_{1}}$ and ${{\mathcal{G}}_{2}}$ by performing data augmentation on $\mathcal{G}$;
      \State Obtain node embeddings ${\*{U}^{k}}\text{=}f\left( {{\mathcal{G}}_{1}} \right)$;
      \State Obtain node embeddings ${\*{V}^{k}}\text{=}f({{\mathcal{G}}_{2}})$ ;
      \State ${\*{U}^{k}}$ and ${\*{V}^{k}}$ are mapped into the contrast space to get $\*{Z}_{u}^{k}$ and $\*{Z}_{v}^{k}$;
      \State Update parameters by applying gradient descent to maximize Eq. \ref{equ:6}.
    \EndFor
  \end{algorithmic}
\end{spacing}
\end{algorithm}

\section{Experiments and Analysis}
\label{sec:10}
\subsection{Dataset Description}
\label{sec:11}
We conducted experiments on eight widely used datasets to compared AMC-GNN with previous graph contrastive learning methods. Cora, Citeseer \cite{article7}, Pubmed \cite{article6} are widely used citation networks where each node represents an article and the edges indicate the citation relationships between articles. DBLP is a co-authorship multi-dimensional graph-based on publication records recorded in computer science literature websites \cite{article35}. We divide the training set, validation set, and test set as its original literature for the above datasets. Coauthor-CS and Coauthor-Physics \cite{article8} are two co-authorship graphs, where nodes are authors connected by their co-authorship. Amazon-Computers and Amazon-Photo \cite{article8} are two networks of co-buy relationships. For the above datasets, we randomly select 10\% of the nodes as the training set, 10\% as the validation set, and the remaining nodes as the test set. The details of each dataset are shown in Table \ref{tab:1}.

\begin{table}[h]\small
\centering
\caption{Statistics of datasets.}
\label{tab:1}       
\begin{tabular}{ccccc}
\hline\noalign{\smallskip}
Dataset	&Nodes &Edges &Features &Classes\\
\noalign{\smallskip}\hline\noalign{\smallskip}
Cora        &2,708 &5,429 &1,433 &7\\
Citeseer	&3,327 &4,732 &3,703 &6\\
Pubmed	     &19,717 &44,338 &500 &3\\
DBLP	     &17,716 &105,734 &1,639 &4\\
Coauthor-CS	 &18,333 &81,894 &6,805	 &15\\
Coauthor-Physics   &34,493 &247,962 &8,415 &5\\
Amazon-Computers   &13,752 &245,861	&767   &10\\
Amazon-Photo       &7,650  &119,081	&745   &8\\
\noalign{\smallskip}\hline
\end{tabular}
\end{table}

\subsection{Experiment Setups}
\label{sec:12}
\paragraph{Transductive Learning}~{}
\newline
In transductive learning tasks, the training set is $D=\left\{ \*{X},{\*{y}_{tra}} \right\}$ and the test sample ${\*{X}_{test}}$ also appears in the training set. We compared two typical transductive learning GNN models, GCN \cite{article15} and SCG \cite{article34}. In the transductive learning task, our encoder can be represented as:

\begin{equation}
\label{equ:9}
{\*{H}^{(l+1)}}=\sigma ({{\tilde{\*{D}}}^{-\frac{1}{2}}}\tilde{\*{A}}{{\tilde{\*{D}}}^{-\frac{1}{2}}}{\*{H}^{l}}{\*{W}^{l}}),
\end{equation}

\noindent where $\tilde{\*{A}}$ is the symmetric normalized adjacency matrix, $\sigma (\cdot )$ denotes the activation function, $\tilde{\*{D}}$ is the diagonal degree matrix ${{\tilde{D}}_{ii}}=\sum\nolimits_{j}{{{{\tilde{A}}}_{ij}}}$, and ${\*{W}^{l}}$ is the trainable weight matrix and is ${\*{H}^{l}}$ the hidden node representation in the $l$th layer. In SCG, The activation function is omitted.

\paragraph{Inductive Learning}~{}
\newline
In inductive learning tasks, the training set is $D=\left\{ {\*{X}_{tra}}, {\*{y}_{tra}} \right\}$ and the test set ${\*{X}_{test}}$ does not appear in the training set. Compared with transductive learning, inductive learning is more flexible. Inductive learning can easily get the representation of a new node by learning a method of node representation, instead of a fixed representation of a node.
Typical inductive learning GNN models are GraphSage \cite{article22} and GAT \cite{article21}, where the model consists of two phases: sampling and aggregation. In the sampling phase, a certain number of neighboring vertices are sampled for each vertex using the connectivity information. In the aggregation phase, the information of neighboring nodes is continuously merged by a multi-layer aggregation function. The merged information is used to predict the node labels. The propagation of the $k$th layer is represented as Eq. \ref{equ:10} and Eq. \ref{equ:11}.

\begin{equation}
\label{equ:10}
\*{h}_{\mathcal{N}(v)}^{k}\leftarrow AGGREGAT{{E}_{k}}\left( \left\{ \*{h}_{u}^{k-1},\forall u\in \mathcal{N}(v) \right\} \right),
\end{equation}

\begin{equation}
\label{equ:11}
\*{h}_{v}^{k}\leftarrow \sigma \left( {{\*{W}}^{k}}\cdot CONCAT(\*{h}_{v}^{k-1},\*{h}_{\mathcal{N}(v)}^{k}) \right),
\end{equation}

\noindent where $\*{h}_{\mathcal{N}(v)}^{k}$ is the embedding of the vertex $v$ at the $k$th layer with ${{\*{h}}^{0}}=\*{x}$, $\mathcal{N}(v)$ is a set of vertices adjacent to $v$.

\paragraph{Baseline}~{}
\newline
Following \cite{article32}, we conduct the node classification task to make the comparison with different methods. In the baseline model, we use traditional methods including: node2vec \cite{article14}, DeepWalk \cite{article19}; We also use Graph Autoencoders (GAE, VGAE) \cite{article33} and graph contrastive learning methods that currently reach state of the art: DGI \cite{article32}, GCA \cite{article2}, GraphCL \cite{article10} and GRACE \cite{article12}. We also compare our experimental results with some classical supervised learning models, including GCN \cite{article15}, SCG \cite{article34}, GraphSage \cite{article22}, GAT \cite{article21}.

\paragraph{Parameter Settings}~{}
\newline
We implement our method with Pytorch and PyTorch Geometric \cite{article16}. We train the model for a fixed number of epochs, 1500, 1000 epochs for Pubmed, DBLP respectively, and 200 epochs for the rest datasets. Adam optimizer are used on all datasets, weight decay of 1e-4. The probability control parameters ${{p}_{R,1}}$, ${{p}_{A,1}}$, ${{q}_{R,1}}$, ${{q}_{A,1}}$ for the first view ${{\mathcal{G}}_{1}}$ and ${{p}_{R,2}}$, ${{p}_{A,2}}$, ${{q}_{R,2}}$, ${{q}_{A,2}}$ for the second view ${{\mathcal{G}}_{2}}$, are all selected between 0 and 1. We conduct experiments on a computer server with four NVIDIA Tesla V100S GPUs (24GB memory each). All dataset-specific hyperparameters are summarized in Table \ref{tab:2}.

\begin{table}[h]
\centering
\caption{Hypeparameter specifications.}
\label{tab:2}       
\resizebox{\textwidth}{!}{
\begin{tabular}{ccccccccccccc}
\hline\noalign{\smallskip}
 Dataset & ${{p}_{R,1}}$ & ${{p}_{R,2}}$ & ${{q}_{R,1}}$ & ${{q}_{R,2}}$ & ${{p}_{A,1}}$ & ${{p}_{A,2}}$ & ${{q}_{R,1}}$ & ${{q}_{R,2}}$ &$\tau $ &\tabincell{c}{Learning\\rate} & \tabincell{c}{Hidden
\\dimension} & \tabincell{c}{Activation\\function}\\
\noalign{\smallskip}\hline\noalign{\smallskip}
Cora     &0.3&0.4&0.2&0.7&0.3&0.3&0.1&0&0.9&5e-4&128 &ReLu  \\
Citeseer &0.3&0.2&0.5&0.5&0.1&0.4&0.1&0.1&0.9&1e-3&256 &PReLu \\
Pubmed   &0.3&0.2&0.6&1.0&0.3&0&0.1&0.1&0.5&1e-3&256 &Relu  \\
DBLP    &0.3&0.2&0.3&0.3&0.2&0.3&0.2&0.2&0.45&1e-3&256 &Relu  \\
Coauthor-CS  &0.3&0.3&0.6&0.3&0.3&0.2&0.1&0&0.8&5e-4&128 &Relu\\
Coauthor-Physics&0.3&0.3&0.2&0.7&0.3&0.2&0.1&0&0.4&1e-3&256&Relu \\
Amazon-Computers&0.3&0.5&0.8&0.2&0.2&0.2&0.1&0&0.7&1e-3&256&Relu \\
Amazon-Photo&0.3&0.2&0.5&0.8&0.2&0.2&0.1&1&0.9&1e-3&256&Relu\\
\noalign{\smallskip}\hline
\end{tabular}}
\end{table}

\paragraph{Results}~{}
\newline
The experimental results are summarized in Table \ref{tab:3}, Training Data denotes the data required for method training. $\*{X}$, $\*{A}$, $\*{Y}$ represent the feature matrix, adjacency matrix and sample labels of the graph, respectively. We use the average after five runs with different random seeds as the results.

\begin{table}[h]
\centering
\caption{Table of experimental results. For transductive tasks and inductive tasks, we use accuracy in percentage and micro-averaged F1-score as results respectively. For clarity, the best performance of AMC-GNN is highlighted in boldface, and the best representations of other unsupervised and supervised graph models are underlined.}
\label{tab:3}       
\resizebox{\textwidth}{!}{
\begin{tabular}{cccccccccc}
\hline\noalign{\smallskip}
Method & \tabincell{c}{Training\\Data} & Cora & Citeseer & Pubmed & DBLP &\tabincell{c}{Coauthor\\-CS} &\tabincell{c}{Coauthor\\-Physics} & \tabincell{c}{Amazon\\-Computers} & \tabincell{c}{Amazon\\-Photo}\\
\noalign{\smallskip}\hline\noalign{\smallskip}
Raw features          &$X$ &64.8 &64.6 &84.8 &71.6 &90.4 &93.6 &73.8 &78.5\\
node2vec              &$A$ &74.8 &52.3 &80.3 &78.8 &85.1 &91.2 &84.4 &89.7\\
DeepWalk              &$A$ &75.7 &50.5 &80.5 &75.9 &84.6 &91.8 &85.7 &89.4\\
DeepWalk + features   &$X,A$ &73.1 &47.6 &83.7 &78.1 &87.7 &94.9 &86.3 &90.1\\
GAE                   &$X,A$ &76.9 &60.6 &82.9 &81.2 &91.6 &94.9 &85.3 &91.6\\
VGAE                  &$X,A$ &78.9 &61.2 &83.0 &81.7 &92.1 &94.5 &86.4 &92.2\\
DGI                &$X,A$ &82.6$\pm$0.4 &68.8$\pm$0.7 &86.0$\pm$0.1 &83.2$\pm$0.1 &\underline{91.4$\pm$0.1} &94.5$\pm$0.5 &84.0$\pm$0.3 &91.6$\pm$0.2\\
GCA                &$X,A$ &83.3$\pm$0.3 &69.3$\pm$0.4 &85.7$\pm$0.1 &83.6$\pm$0.2 &91.1$\pm$0.1 &93.7$\pm$0.3 &84.3$\pm$0.2 &90.6$\pm$0.2\\
GraphCL             &$X,A$ &\underline{83.5$\pm$0.3} &71.2$\pm$0.5 &84.6$\pm$0.1 &\underline{84.5$\pm$0.1} &91.1$\pm$0.1 &93.2$\pm$0.3 &85.3$\pm$0.2 &90.8$\pm$0.2\\
GRACE              &$X,A$ &83.3$\pm$0.4 &\underline{72.1$\pm$0.5} &\underline{86.7}$\pm$0.1 &84.2$\pm$0.1 &89.8$\pm$0.3 &\underline{95.6$\pm$0.2} &\underline{87.3$\pm$0.2} &\underline{92.1$\pm$0.2}\\
\noalign{\smallskip}\hline\noalign{\smallskip}
AMC-GCN           &$X,A$ &\textbf{84.8$\pm$0.4} &\textbf{72.8$\pm$0.5} &\textbf{87.1$\pm$0.1} &\textbf{84.9$\pm$0.2} &\textbf{92.4$\pm$0.2} &95.7$\pm$0.1 &88.9$\pm$0.2 &92.8$\pm$0.1\\
AMC-GAT           &$X,A$ &84.6$\pm$0.3 &72.3$\pm$0.3 &86.9$\pm$0.2 &84.3$\pm$0.1 &91.5$\pm$0.1 &\textbf{95.9$\pm$0.1} &\textbf{89.5$\pm$0.2} &\textbf{93.1$\pm$0.2}\\
\noalign{\smallskip}\hline\noalign{\smallskip}
GCN                &$X,A,Y$ &82.8 &72.0 &\underline{84.9} &82.7 &\underline{93.0} &\underline{95.7} &86.5 &92.4\\
SGC                &$X,A,Y$ &80.6 &69.1 &84.8 &81.7 &92.1 &95.1	&83.9 &90.9\\
GAT                &$X,A,Y$ &\underline{83.7} &\underline{72.5} &79.3 &\underline{83.7} &92.3 &95.5 &\underline{86.9} &\underline{92.6}\\
\noalign{\smallskip}\hline
\end{tabular}}
\end{table}

\begin{figure}[h]
\centering
\subfloat[Raw]{
\begin{minipage}[h]{0.24\linewidth}
\centering
\includegraphics[width=1.5in]{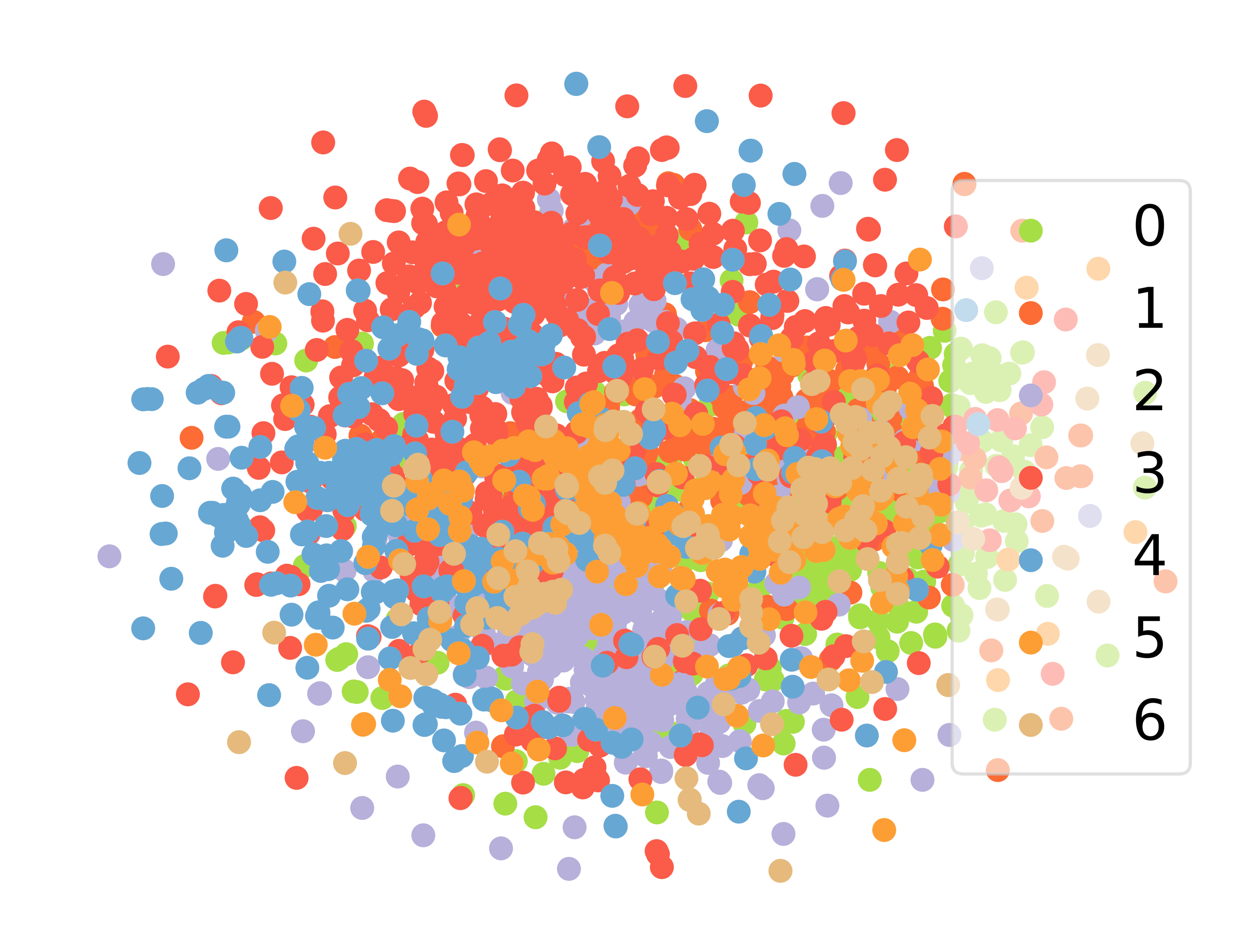}
\end{minipage}%
}%
\subfloat[DGI]{
\begin{minipage}[h]{0.24\linewidth}
\centering
\includegraphics[width=1.5in]{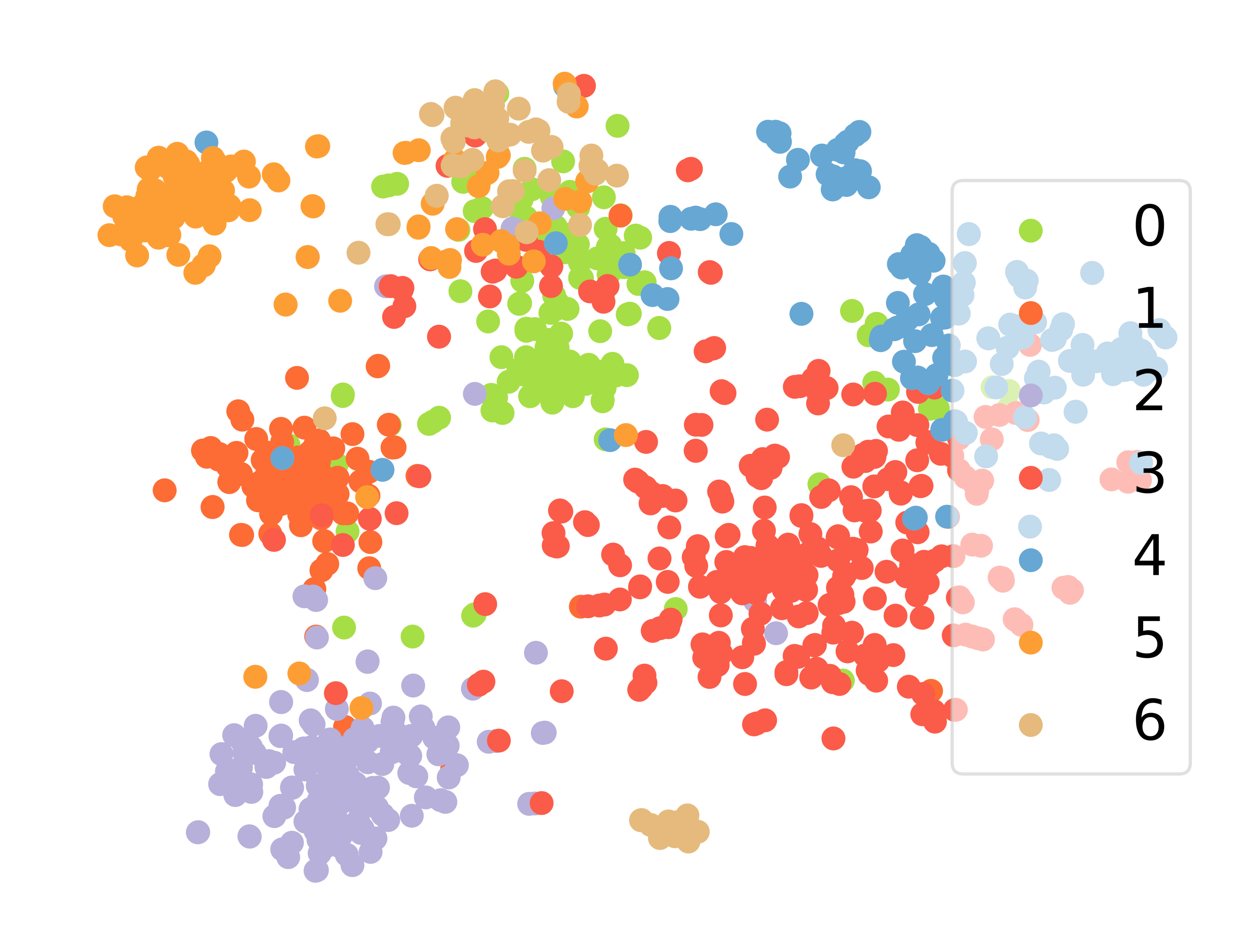}
\end{minipage}%
}%
\subfloat [GRACE]{
\begin{minipage}[h]{0.24\linewidth}
\centering
\includegraphics[width=1.5in]{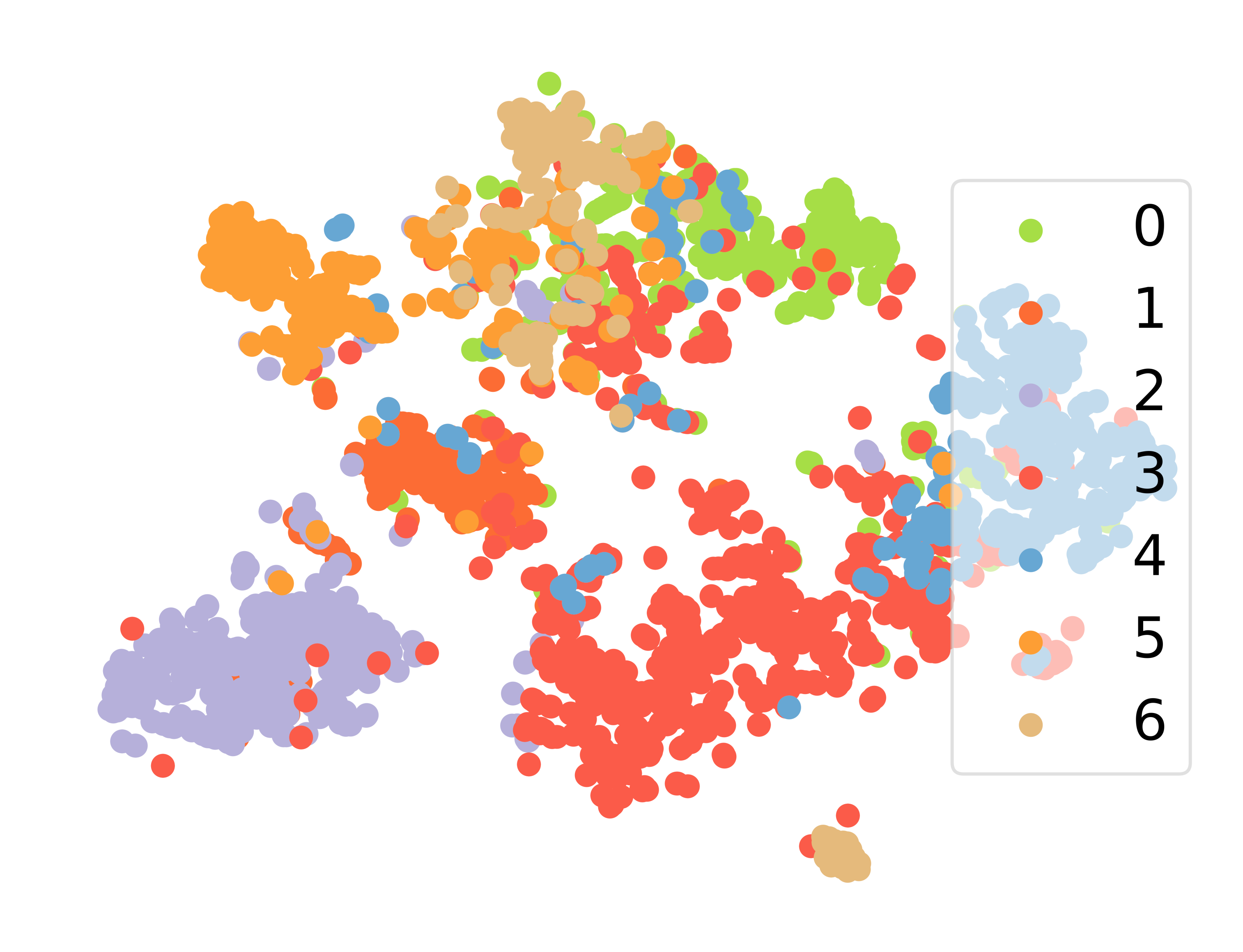}
\end{minipage}%
}%
\subfloat [AMC-GCN]{
\begin{minipage}[h]{0.24\linewidth}
\centering
\includegraphics[width=1.5in]{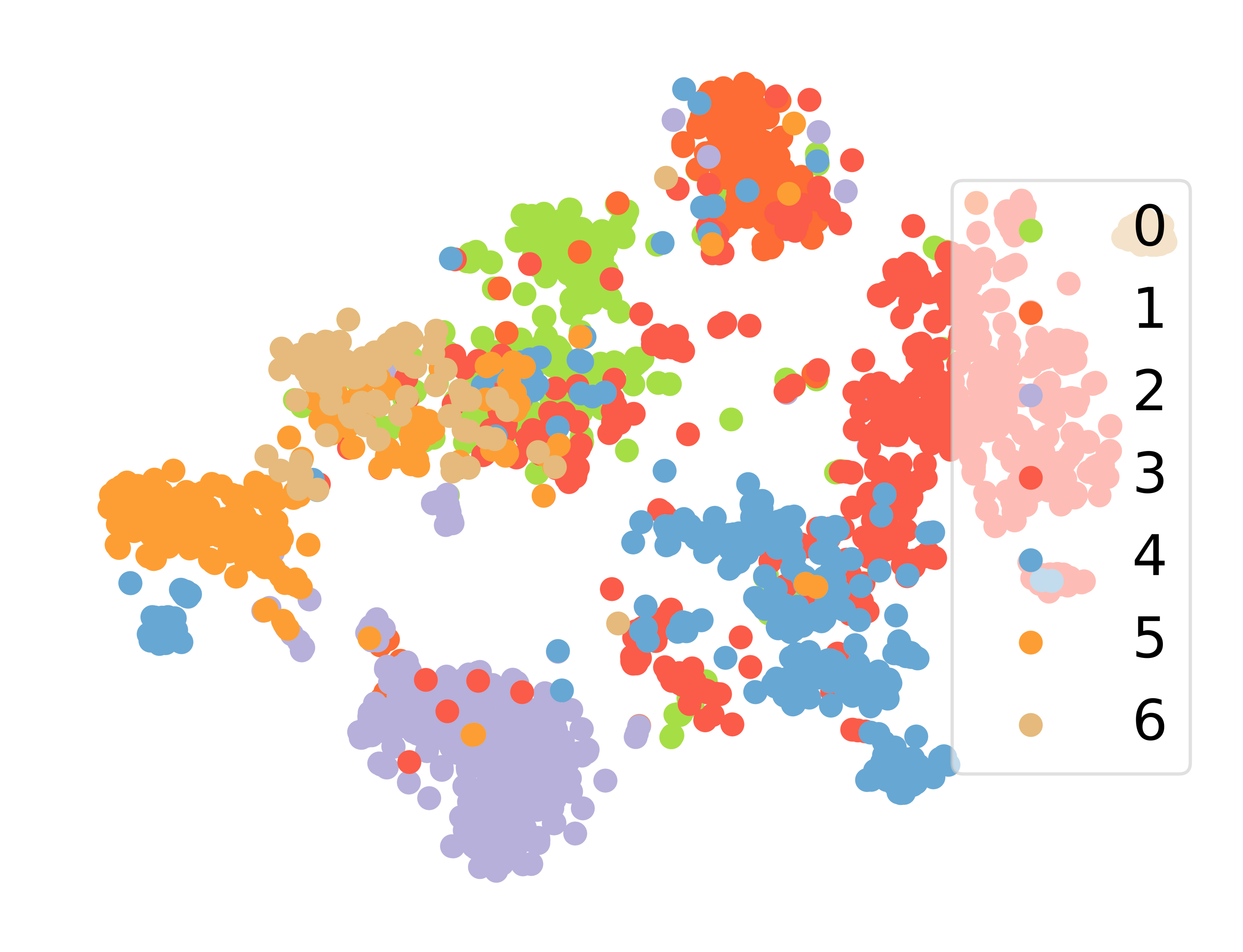}
\end{minipage}%
}%
\caption{The t-SNE visualization of learned node representations on Cora. (a) Raw means the raw node features are used. (b) DGI indicates the features are learned by DGI. (c) GRACE indicates the features are learned by GRACE. (d) AMC-GCN indicates the features are learned by AMC-GCN. The Silhouette score for (a) (b) (c) (d) respectively is 0.005, 0.207, 0.156 and 0.243.}
\label{fig:3}
\centering
\end{figure}

AMC-GCN and AMC-GAT represent AMC-GNN models with GCN and GAT as encoders, respectively. AMC-GCN, GCN, SGC are transductive learning methods. AMC-GAT, GAT are inductive learning methods. The proposed model AMC-GNN outperforms the state-of-the-art unsupervised models DGI , GraphCL, GCA, and GRACE on all the datasets used in the experiments, and the performance on some datasets even exceeds that of supervised learning methods. On the Cora dataset, AMC-GCN outperformed GraphCL by 1.3\% and outperformed DGI by 2.2\% under 200 epochs of training. On the Amazon-Computers dataset, AMC-GAT outperformed GRACE by 2.2\% and outperformed GCA by 5.2\% under 200 epochs of training. We visualized the features after model encoding, using different colored points to represent different classes of samples. We performed the analysis on the Cora dataset because it has the least number of nodes for clear presentation. T-SNE plots of the embeddings is given in Fig. \ref{fig:3}.

\subsection{Ablation Studies}
\label{sec:13}

The experimental results in Table \ref{tab:3} demonstrate the effectiveness of adaptive multi-layer contrastive loss and the auxiliary training model in improving the model's performance. In this section, we will explore the implications of the adaptive multi-layer contrastive loss and auxiliary training model.
\par w/o auxiliary model means using the model without auxiliary model. w/o multi-layer represents the model without adaptive multi-layer contrastive loss function. The performance of the above models was tested, and the models used the same parameters as in Sect. \ref{sec:12}. We run five times and average the results for all methods, and the experimental results are shown in Table \ref{tab:4}.

To reflect the change of feature space, we selected the well known K-means \cite{article40} to cluster the vectors generated by the model, and calculate clustering evaluation indexes such as: Calinski-Harabaz Index(CHI) \cite{article39}, Davies-Bouldin Index(DBI) \cite{article23}, Silhouette Coefficient(SC) \cite{article3}. For CHI and SC, larger values indicate better clustering of features; for DBI, smaller values indicate better clustering of features. A high-performance coding model produces features that are closer together within classes, while features between classes are further apart and more clearly bounded.

\begin{table}[h]\scriptsize
\centering
\caption{The performance of AMC-GCN, w/o auxiliary model and w/o multi-layer in transductive node classification on four citation datasets. The best clustering results are highlighted in boldface.}
\label{tab:4}
\begin{tabular}{c|c|cccc}
\hline\noalign{\smallskip}
Dataset &Model &CHI &DBI &SC &Accuracy \\
\noalign{\smallskip}\hline
\multirow{3}*{Cora}     &full method    &\textbf{204.094}   &\textbf{2.117}   &\textbf{0.088}
&\textbf{84.8$\pm$0.4}
\\
                        &w/o auxiliary model   &174.778   &2.287   &0.085 &84.1$\pm$0.4\\
                        &w/o multi-layer   &128.494   &2.835   &0.033 &83.6$\pm$0.3\\   \hline
\multirow{3}*{Citeseer} &full method    &\textbf{163.685}   &\textbf{2.753}   &\textbf{0.018} &\textbf{72.8$\pm$0.5}\\
                        &w/o auxiliary model &174.778   &2.287 &0.085 &72.5$\pm$0.5\\
                        &w/o multi-layer   &128.494   &2.835 &0.033 &72.1$\pm$0.5\\   \hline
\multirow{3}*{Pubmed}   &full method    &\textbf{3152.101}  &\textbf{2.064}   &\textbf{0.070} &\textbf{87.1$\pm$0.1}\\
                        &w/o auxiliary model   &1544.902  &3.008 &0.026 &86.8$\pm$0.1\\
                        &w/o multi-layer   &1361.766  &2.664 &0.034 &86.5$\pm$0.2\\   \hline
\multirow{3}*{DBLP}     &Full Method    &\textbf{959.503}	 &\textbf{3.161}   &\textbf{0.011}
&\textbf{84.9$\pm$0.2}\\
                        &w/o auxiliary model   &586.525	 &4.262 &-0.012 &84.5$\pm$0.2\\
                        &w/o multi-layer   &425.860	 &5.020 &-0.021 &84.2$\pm$0.3\\
\noalign{\smallskip}\hline
\end{tabular}
\end{table}

The experimental results on four data sets prove that introducing the auxiliary model and the adaptive multi-layer contrastive loss can make the embeddings have a better clustering effect. Compared to the previous graph contrastive learning model, in the feature space, the features of the same category are closer to each other, while the features of different categories are farther away. By learning better embeddings, the performance of the model improved. The clustering effect proves that the introduction of adaptive multi-layer contrastive loss is a crucial reason for improved performance.

\subsection{The analysis of Data Augmentation}
\label{sec:14}
Data augmentation is essential in graph contrastive learning, and in our experiments, we use a combination of different data augmentation approaches to generate different graph views. In this section, we will compare different data augmentation methods and analyze the role of data augmentation.

AMC-GCN(RN) and AMC-GCN(AM) indicate the model with Random Sampling Neighbours and Attribute Masking only respectively. The model parameters are set as Sect. \ref{sec:12}, and the performance of the model for node classification on four datasets: Cora, Citeseer, Pubmed and DBLP, is shown in Table \ref{tab:5}. Obviously, the performance of utilizing both RN and AM data augmentation methods is better than using only a single method, and the results demonstrate that data augmentation requires changes in both graph topology and node features to have better results.

\begin{table}[h]\small
\centering
\caption{The performance of model variants along with the original AMC-GCN model, AMC-GCN(RN) and AMC-GCN(AM).}
\label{tab:5}       
\begin{tabular}{ccccc}
\hline\noalign{\smallskip}
Method	&Cora	  &Citeseer  &Pubmed    &DBLP \\
\noalign{\smallskip}\hline\noalign{\smallskip}
AMC-GCN	    &\textbf{84.8$\pm$0.4} &\textbf{72.8$\pm$0.5} &\textbf{87.1$\pm$0.1} &\textbf{84.9$\pm$0.2}\\
AMC-GCN(RN)	&84.5$\pm$0.2 &68.8$\pm$0.4 &85.4$\pm$0.2 &83.8$\pm$0.4\\
AMC-GCN(AM)	&84.6$\pm$0.2 &70.7$\pm$0.3 &86.1$\pm$0.2 &84.4$\pm$0.2\\
\noalign{\smallskip}\hline
\end{tabular}
\end{table}

\subsection{Analysis of Attention Mechanism}
\label{sec:16}
We analyzed the attention distribution and the attention learning trend separately to explore whether the attention values learned by the attention mechanism are meaningful.

\begin{figure}[h]
\centering
\subfloat[Cora]{
\begin{minipage}[h]{0.24\linewidth}
\centering
\includegraphics[width=1.5in]{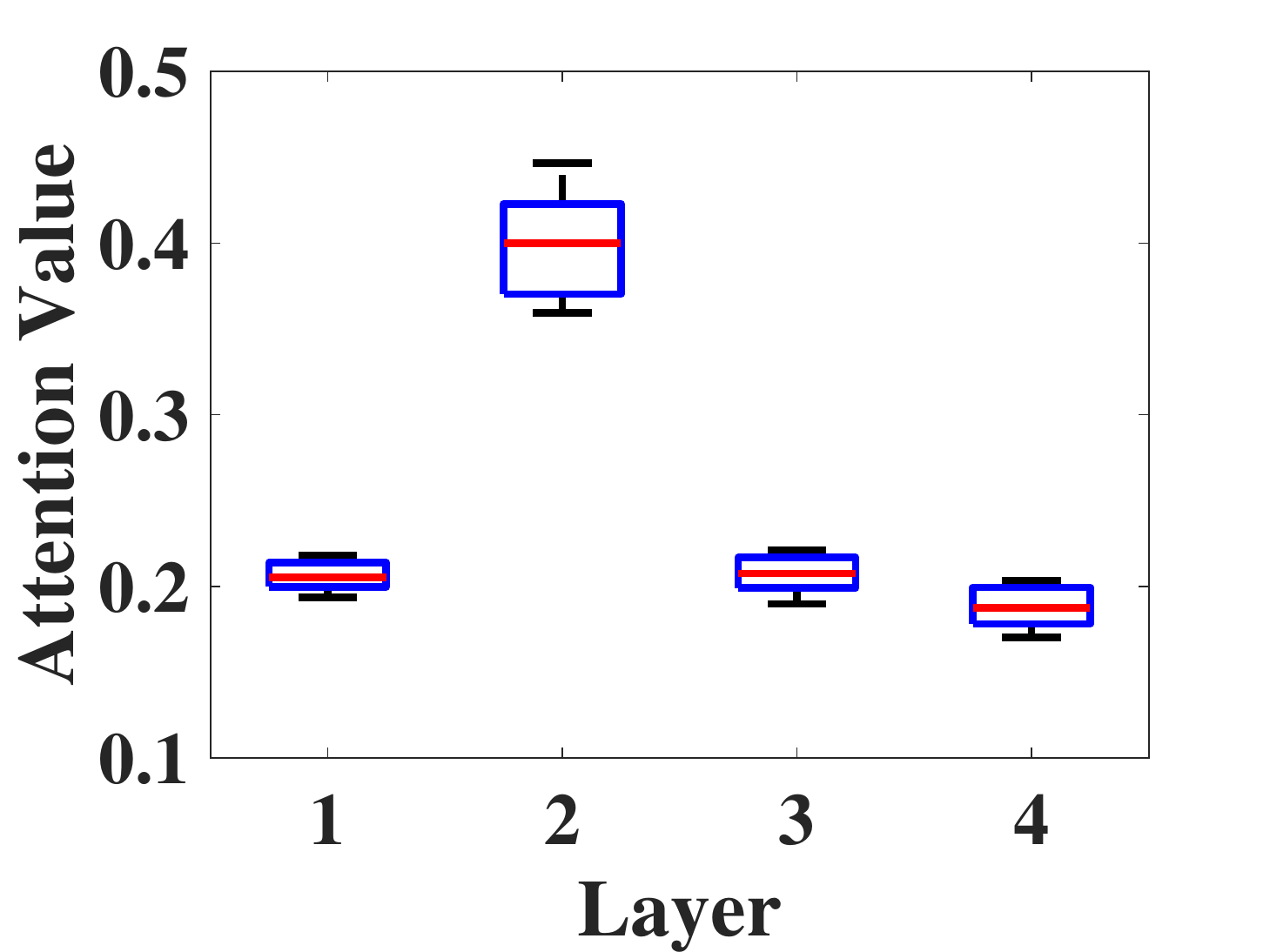}
\end{minipage}%
}%
\subfloat[Citeseer]{
\begin{minipage}[h]{0.24\linewidth}
\centering
\includegraphics[width=1.5in]{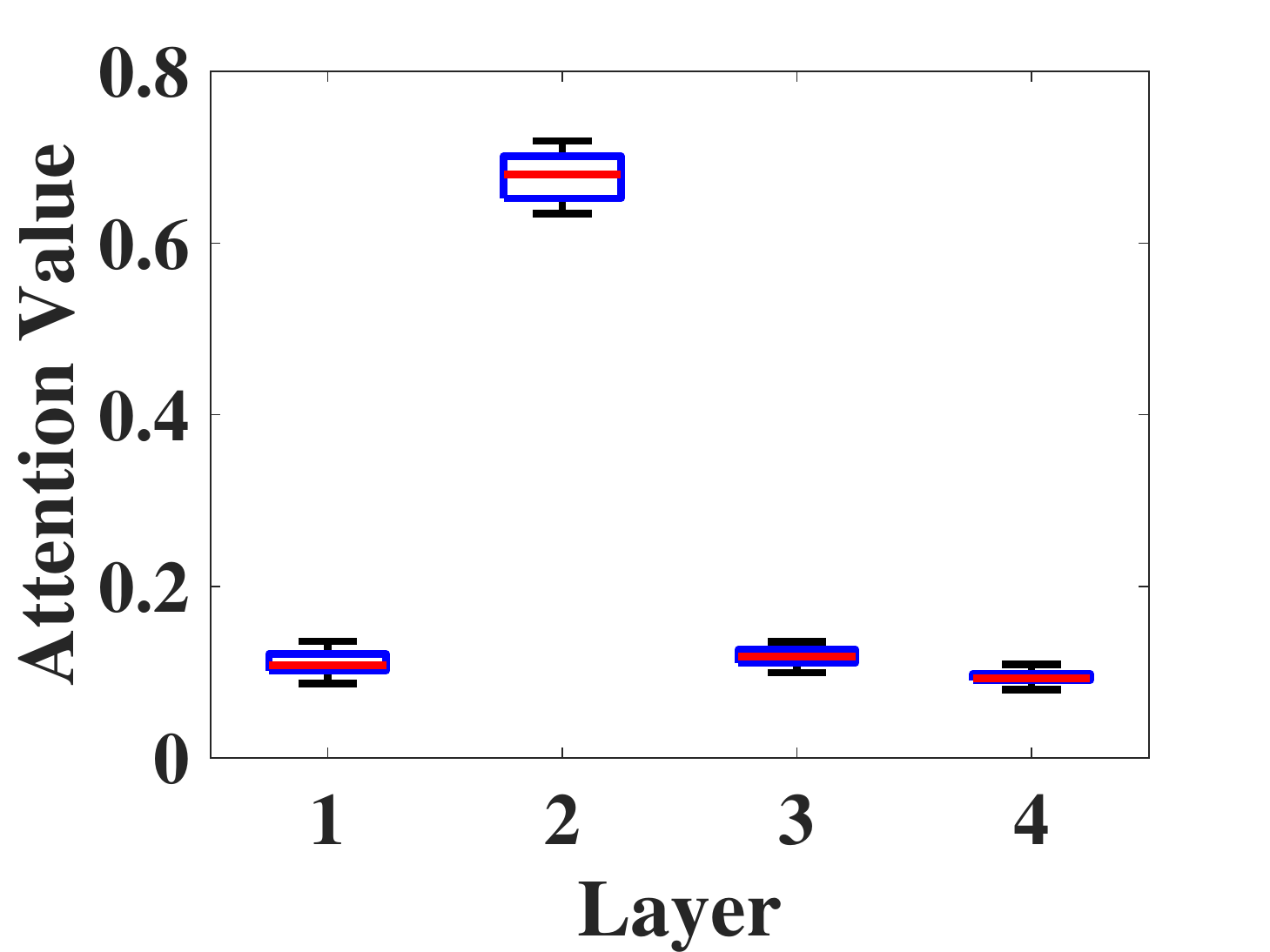}
\end{minipage}%
}
\subfloat [Pubmed]{
\begin{minipage}[h]{0.24\linewidth}
\centering
\includegraphics[width=1.5in]{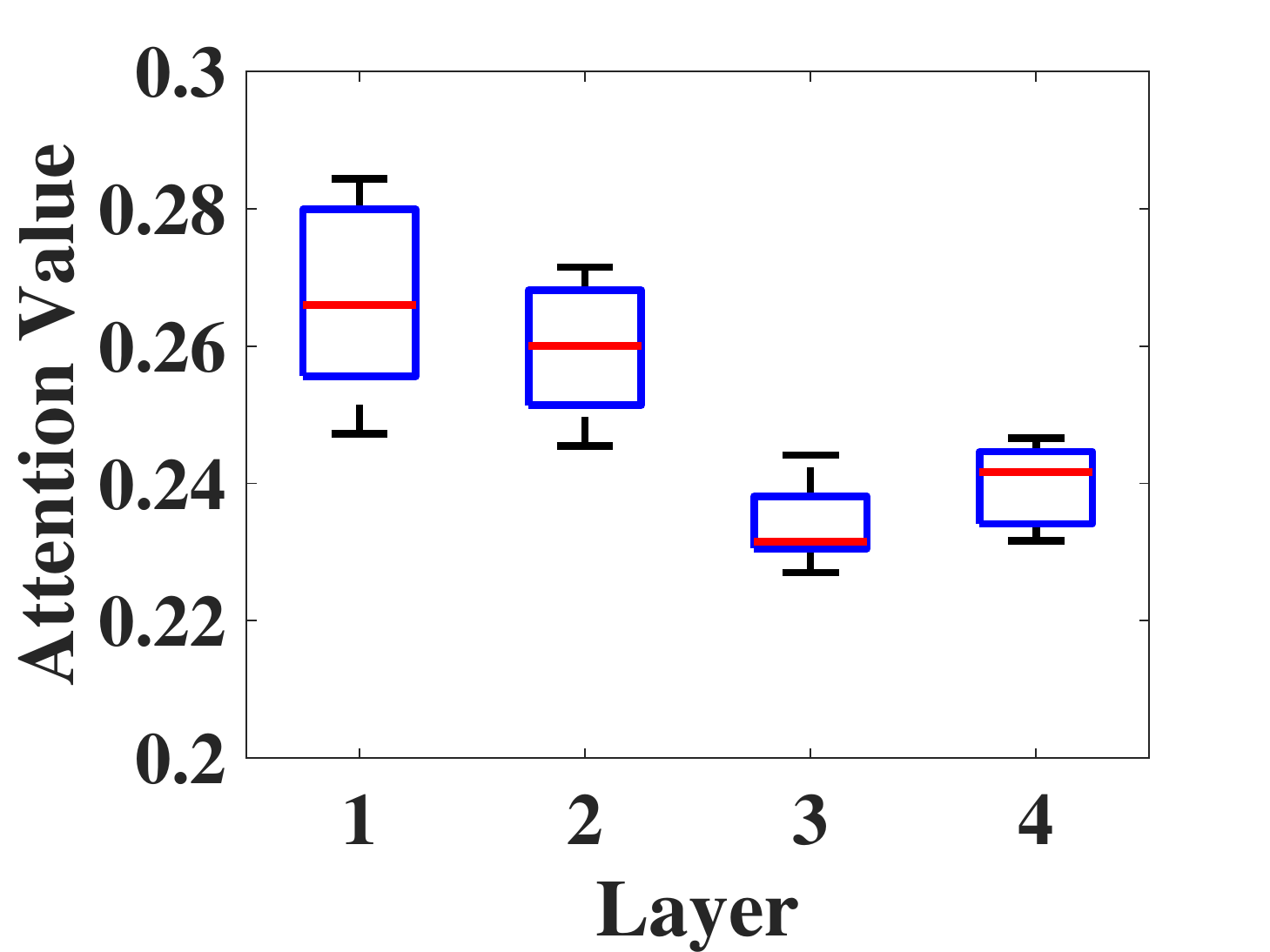}
\end{minipage}%
}%
\subfloat [DBLP]{
\begin{minipage}[h]{0.24\linewidth}
\centering
\includegraphics[width=1.5in]{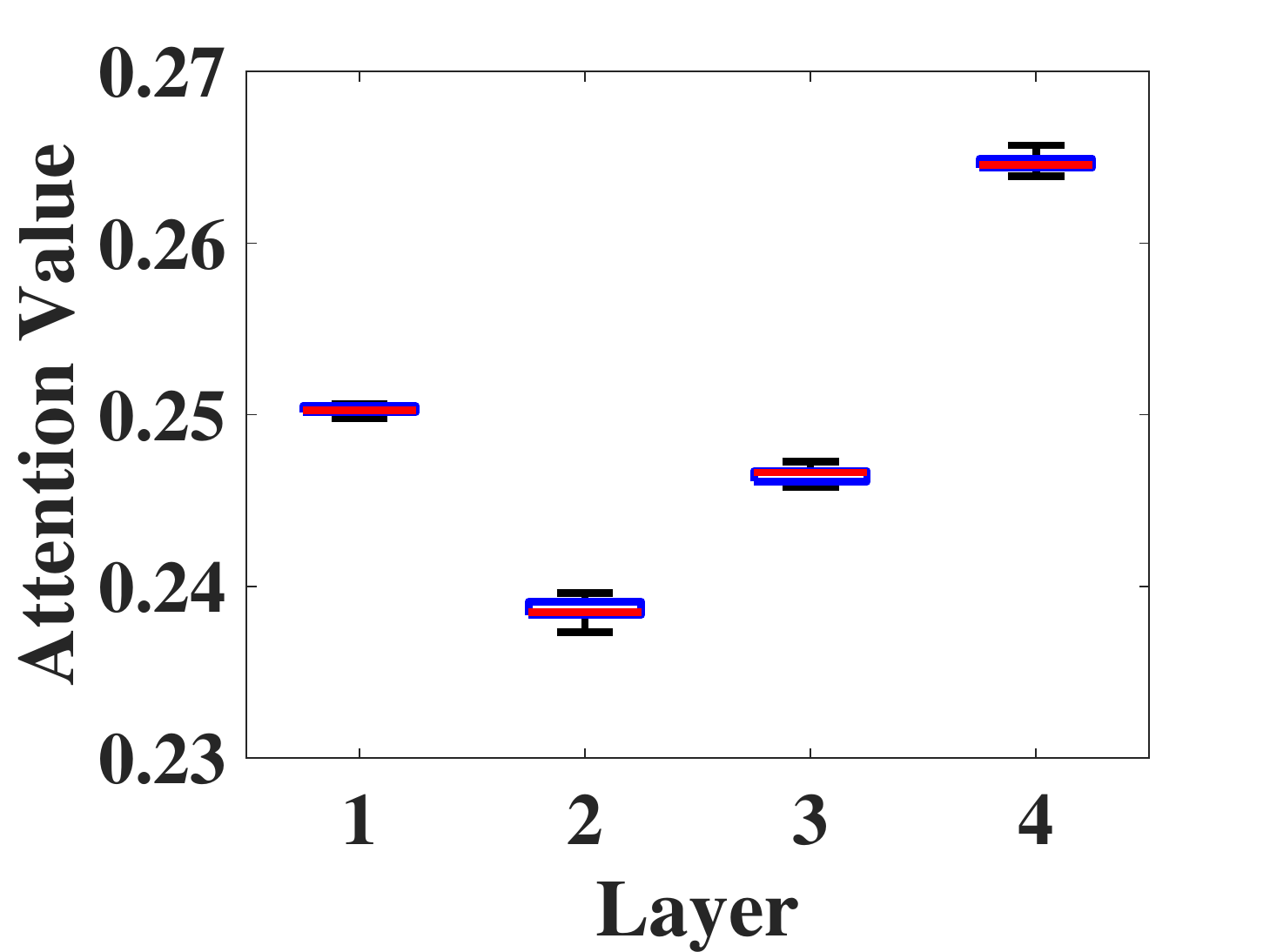}
\end{minipage}%
}%
\caption{Analysis of attention distribution.}
\label{fig:4}
\centering
\end{figure}

\paragraph{Analysis of attention distributions}~{}
\newline
AMC-GNN learns the importance weights of embedding in different layers through the attention mechanism. The training data set and parameter settings are the same as in Sect. \ref{sec:12}. We conduct the attention distribution analysis on four datasets: Cora, Citeseer, Pubmed, and DBLP by using AMC-GCN, where the results are shown in Fig. \ref{fig:4}. As we can see, for Cora, Citeseer, the attention values of explicit embeddings in the second layer are larger than other layers’. This suggests that the data in embeddings of the second layer ought to be more critical than other layers in feature space. For Pubmed, embeddings of first two layers contain more information. For DBLP, embeddings of the last layer are more useful for optimization.

\begin{figure}[h]
\centering
\subfloat[Cora]{
\begin{minipage}[h]{0.24\linewidth}
\centering
\includegraphics[width=1.5in]{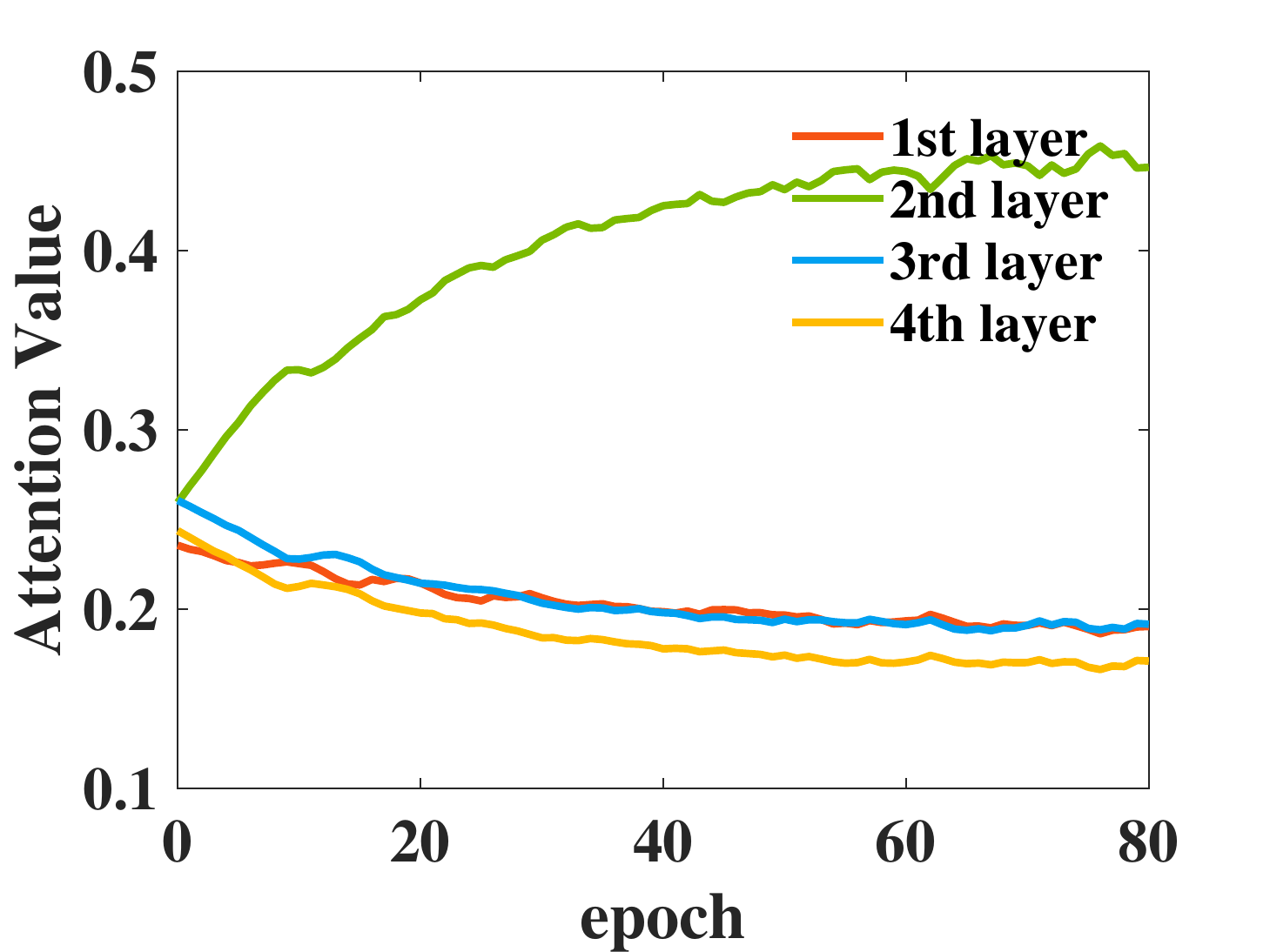}
\end{minipage}%
}%
\subfloat[Citeseer]{
\begin{minipage}[h]{0.24\linewidth}
\centering
\includegraphics[width=1.5in]{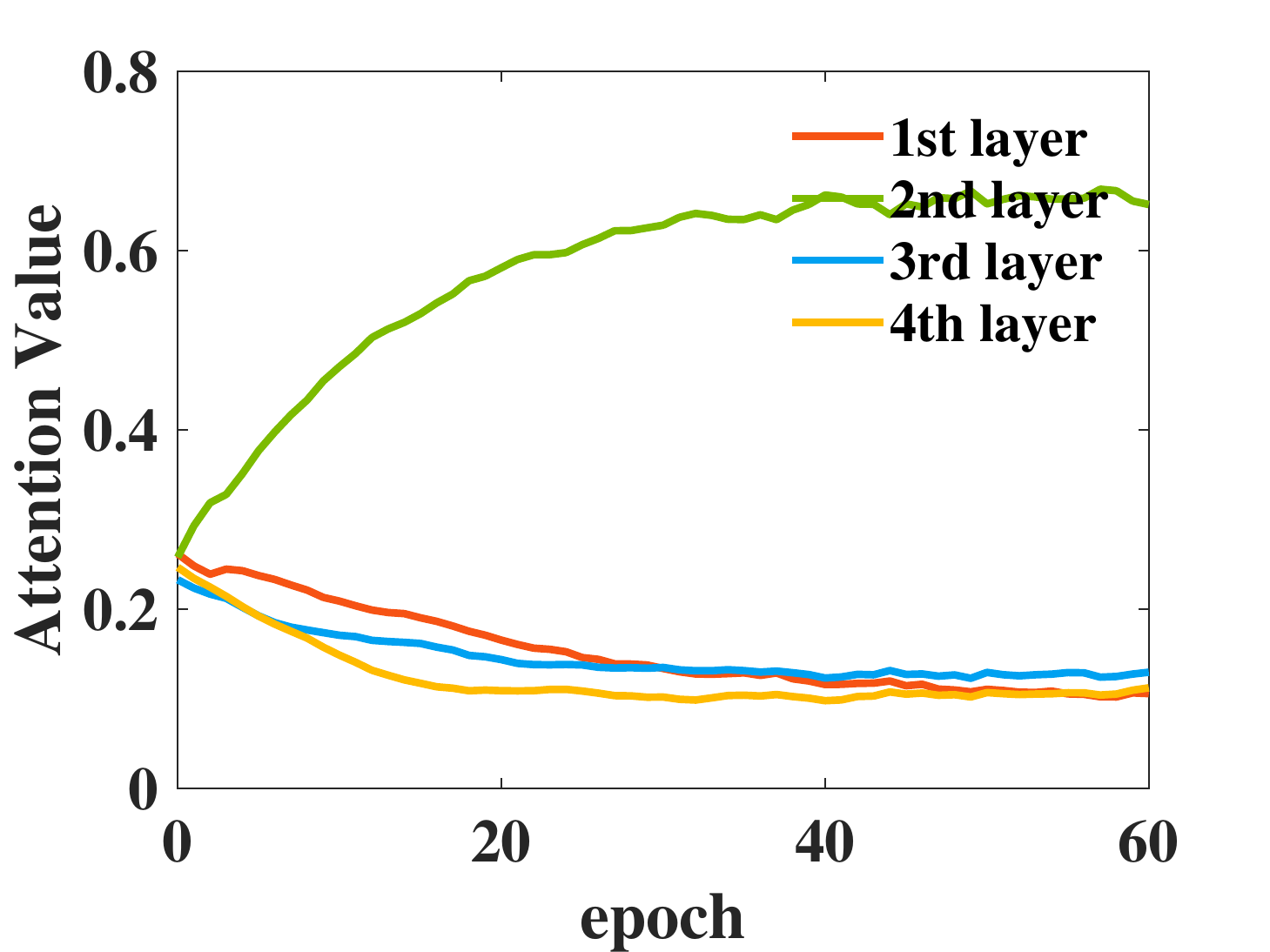}
\end{minipage}%
}
\subfloat[Pubmed]{
\begin{minipage}[h]{0.24\linewidth}
\centering
\includegraphics[width=1.5in]{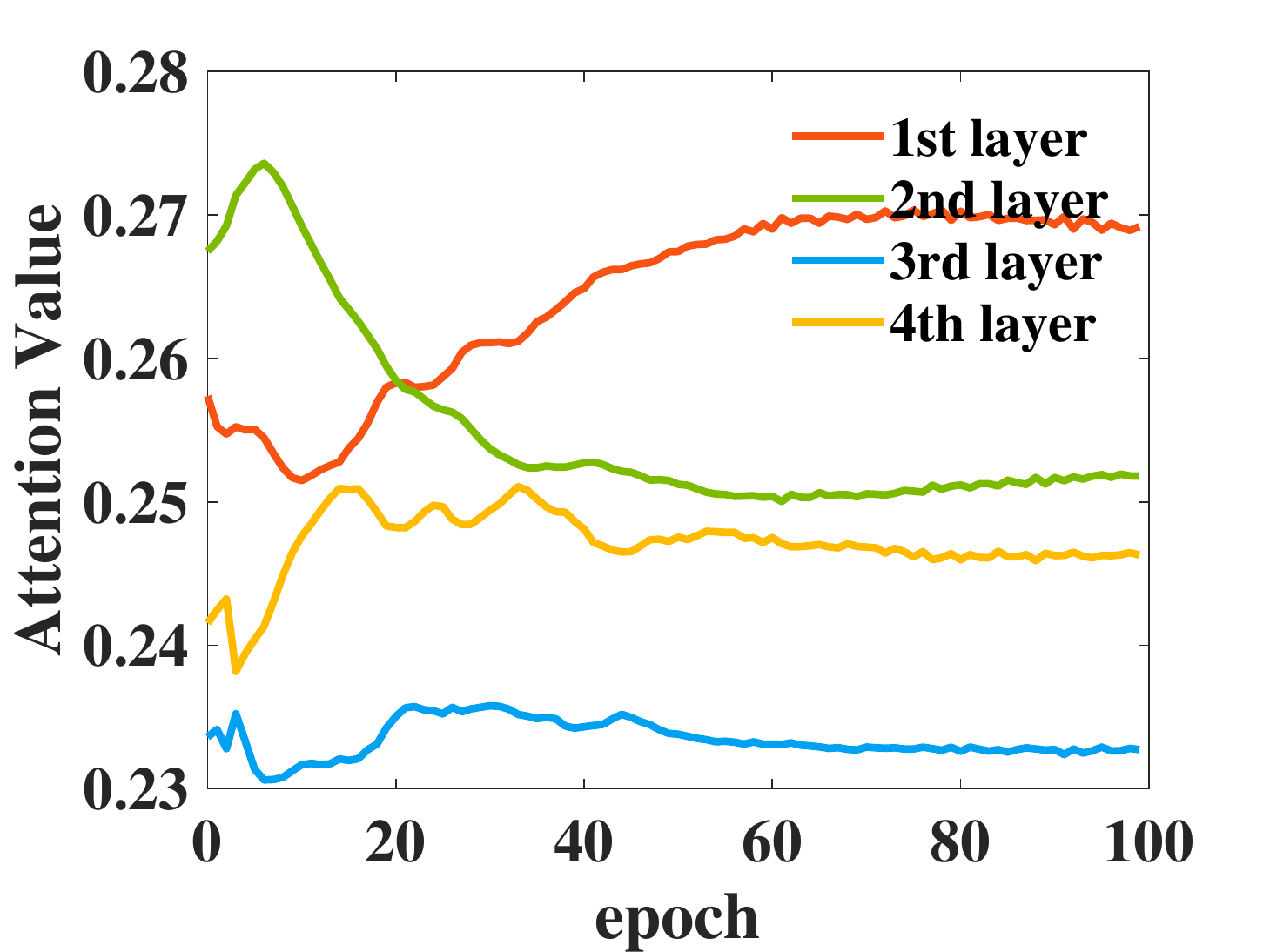}
\end{minipage}%
}%
\subfloat[DBLP]{
\begin{minipage}[h]{0.24\linewidth}
\centering
\includegraphics[width=1.5in]{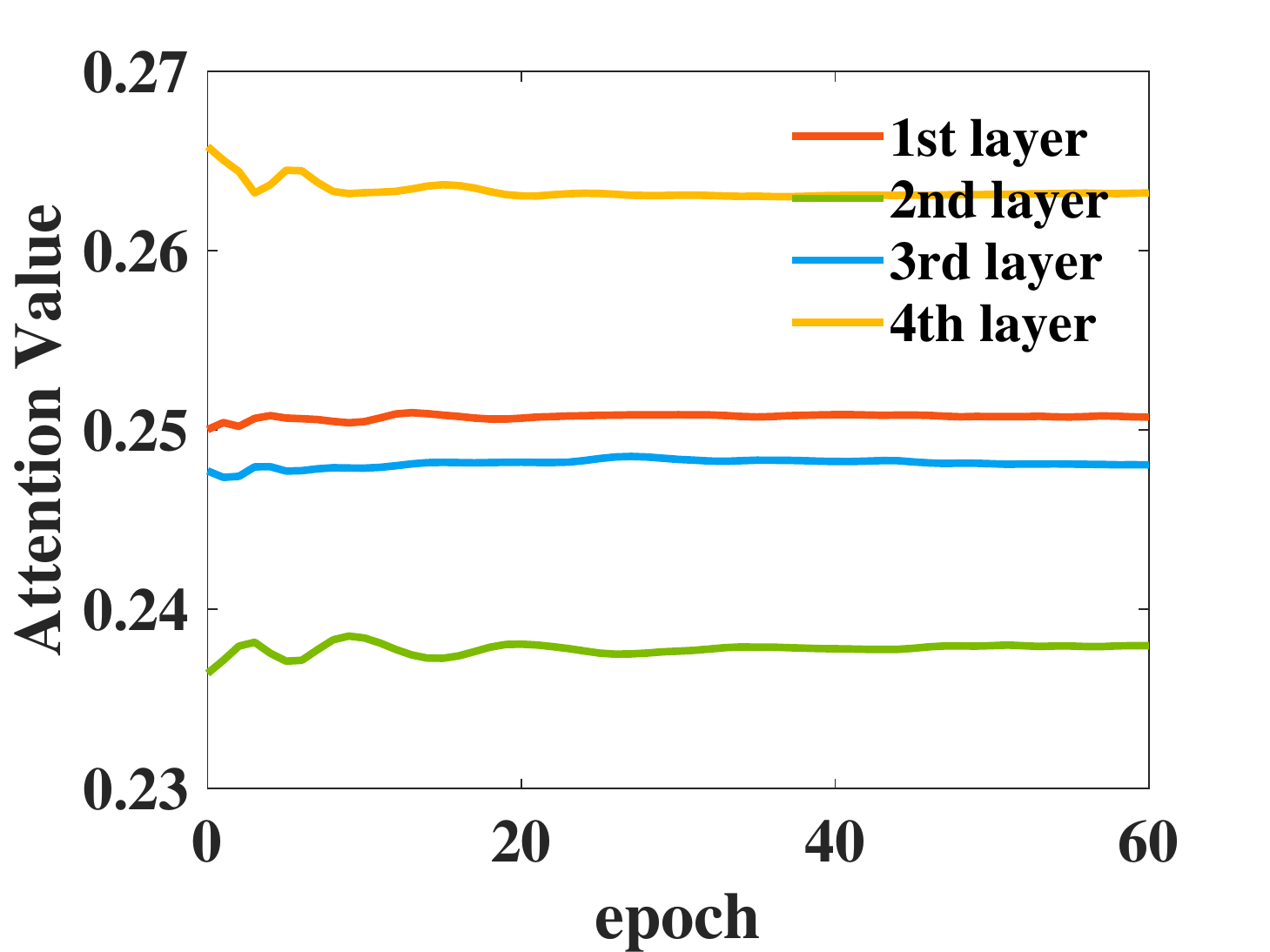}
\end{minipage}%
}%
\caption{The attention changing trends w.r.t epochs.}
\label{fig:5}
\centering
\end{figure}

\paragraph{Analysis of attention trends}~{}
\newline

We dissect the changing patterns of attention value during the preparation cycle. Here we accept Cora and Citeseer as models in Fig. \ref{fig:5}, where x-axis denotes the epoch and y-axis denotes the average attention value. Toward the start, the average attention values of the embeddings of all layers are practically similar, with the training epoch increasing, the attention values become unique. The attention value for embeddings of the second layer gradually increases, while the attention value for other embeddings keeps decreasing. This marvel is reliable with the conclusions in Fig. \ref{fig:4}, and we can see that AMC-GCN can learn the significance of embeddings in different layers gradually.

\subsection{Robustness to Sparse Features}
\label{sec:15}
When a small amount of features are removed, the impact on the model is minor. However, when the node features are too sparse, the nodes may not have enough features to maintain their original semantics and can cause a decrease in the accuracy of the model.
\par In this section, we use AM to randomly pollute the training data and explore the robustness of the model to sparse features. Specifically, we conduct experiments on four cited network datasets, ranging the masking ratio of node features from 0.4 to 0.9. During training, all other parameters of the models are set as Sect. \ref{sec:12}.

\begin{figure}[h]
\centering
\subfloat[Cora]{
\begin{minipage}[h]{0.24\linewidth}
\centering
\includegraphics[width=1.5in]{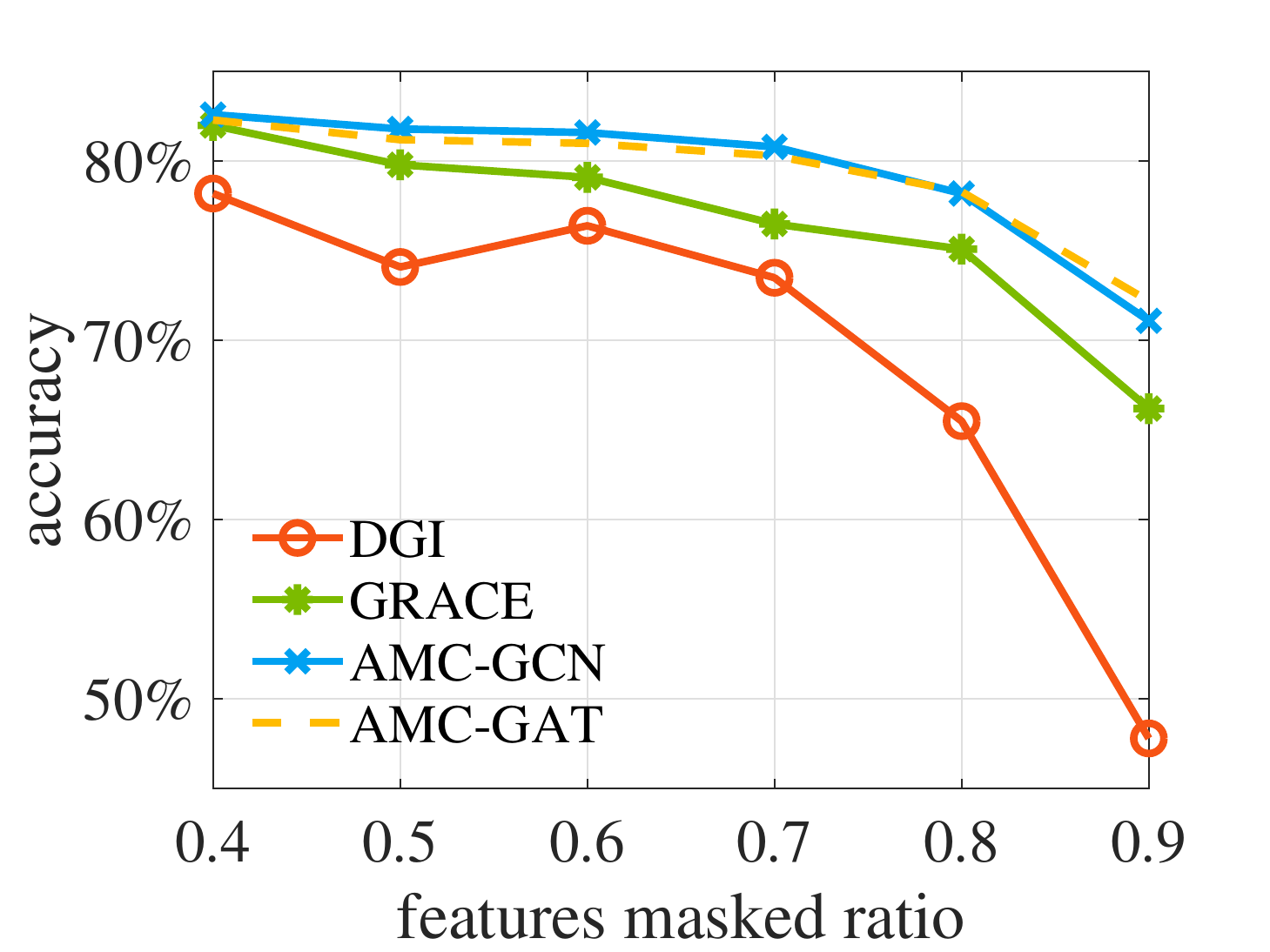}
\end{minipage}%
}%
\subfloat[Citeseer]{
\begin{minipage}[h]{0.24\linewidth}
\centering
\includegraphics[width=1.5in]{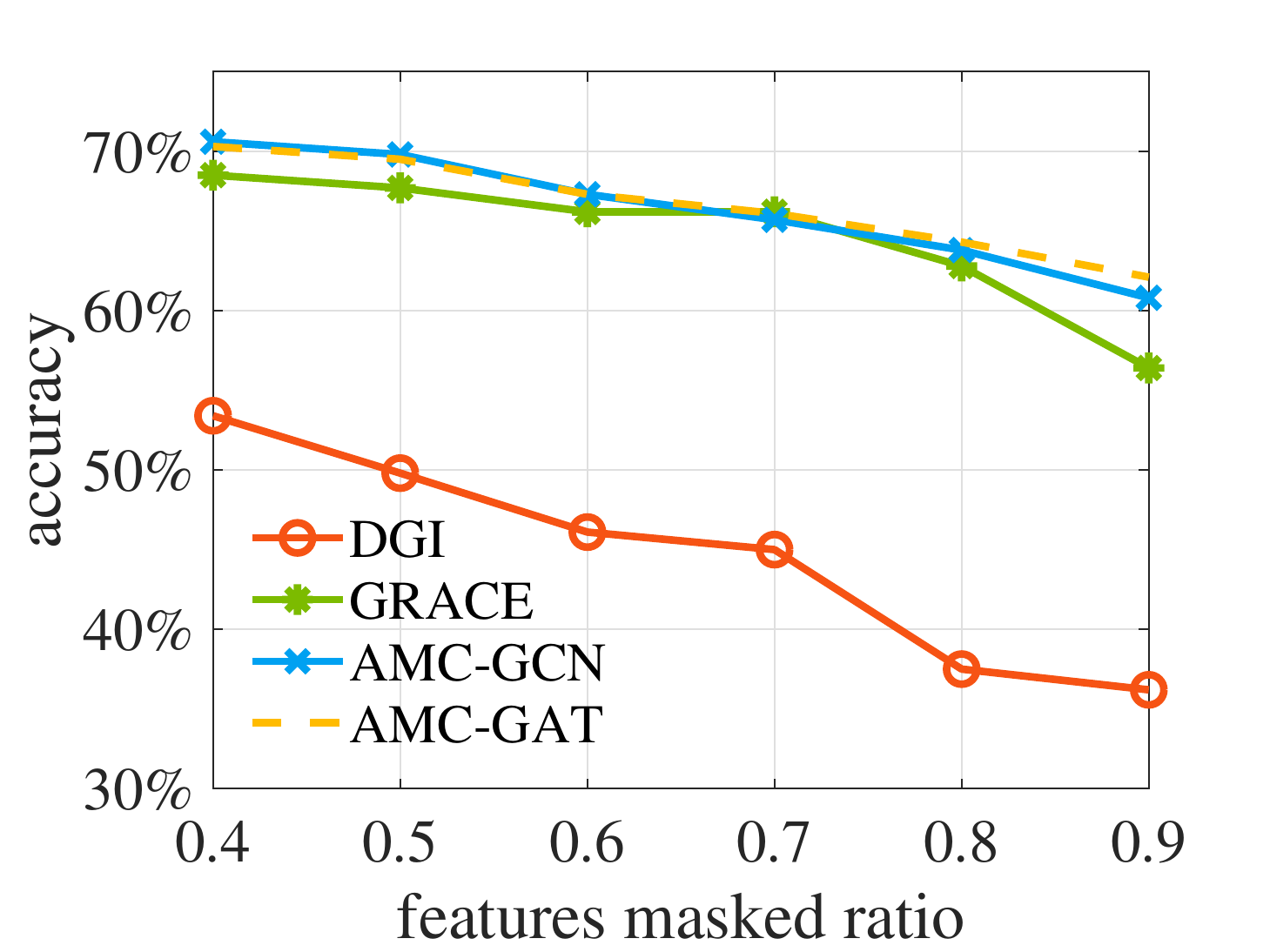}
\end{minipage}%
}
\subfloat [Amazon-Photo]{
\begin{minipage}[h]{0.24\linewidth}
\centering
\includegraphics[width=1.5in]{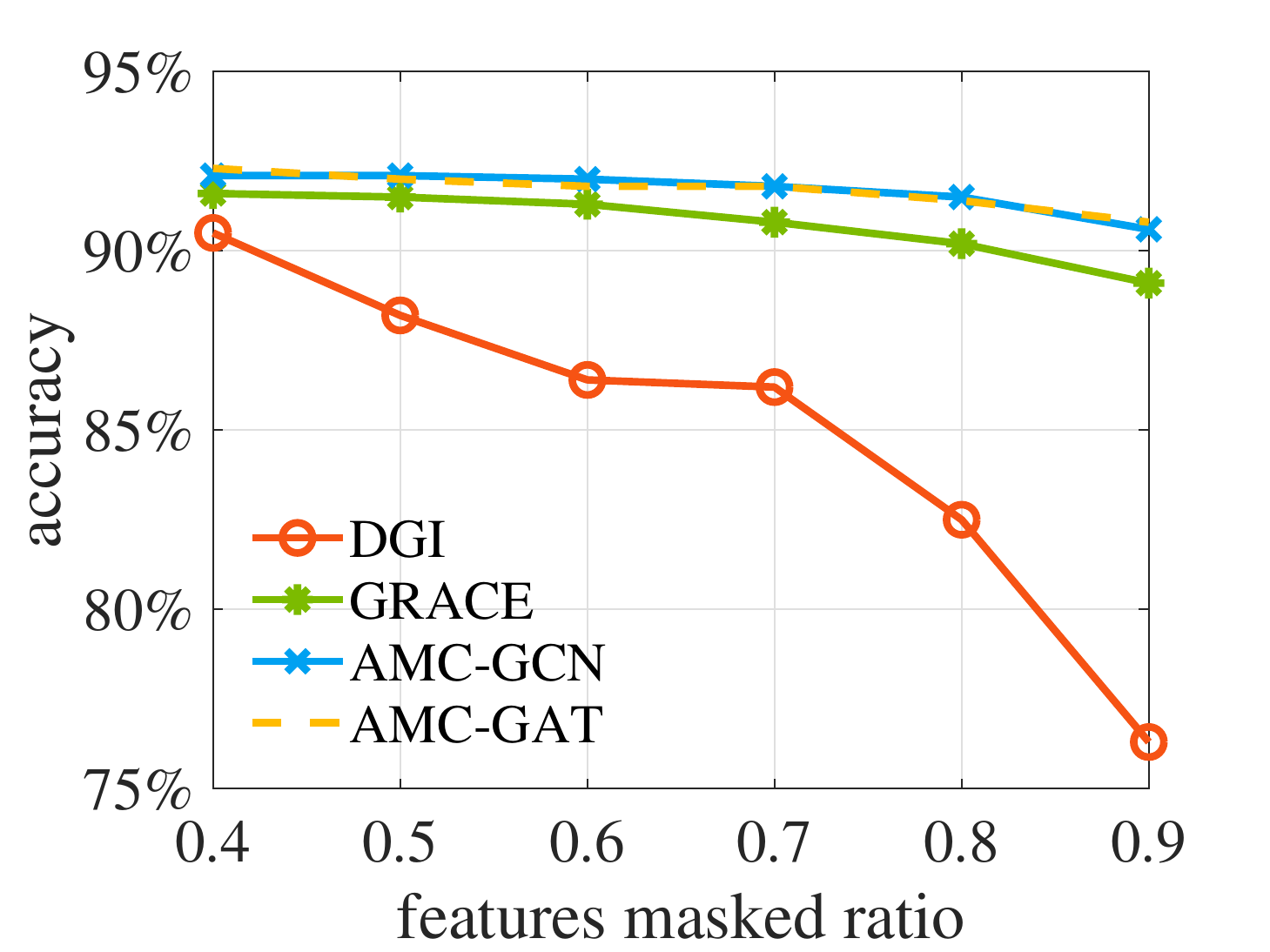}
\end{minipage}%
}%
\subfloat [Amazon-Computer]{
\begin{minipage}[h]{0.24\linewidth}
\centering
\includegraphics[width=1.5in]{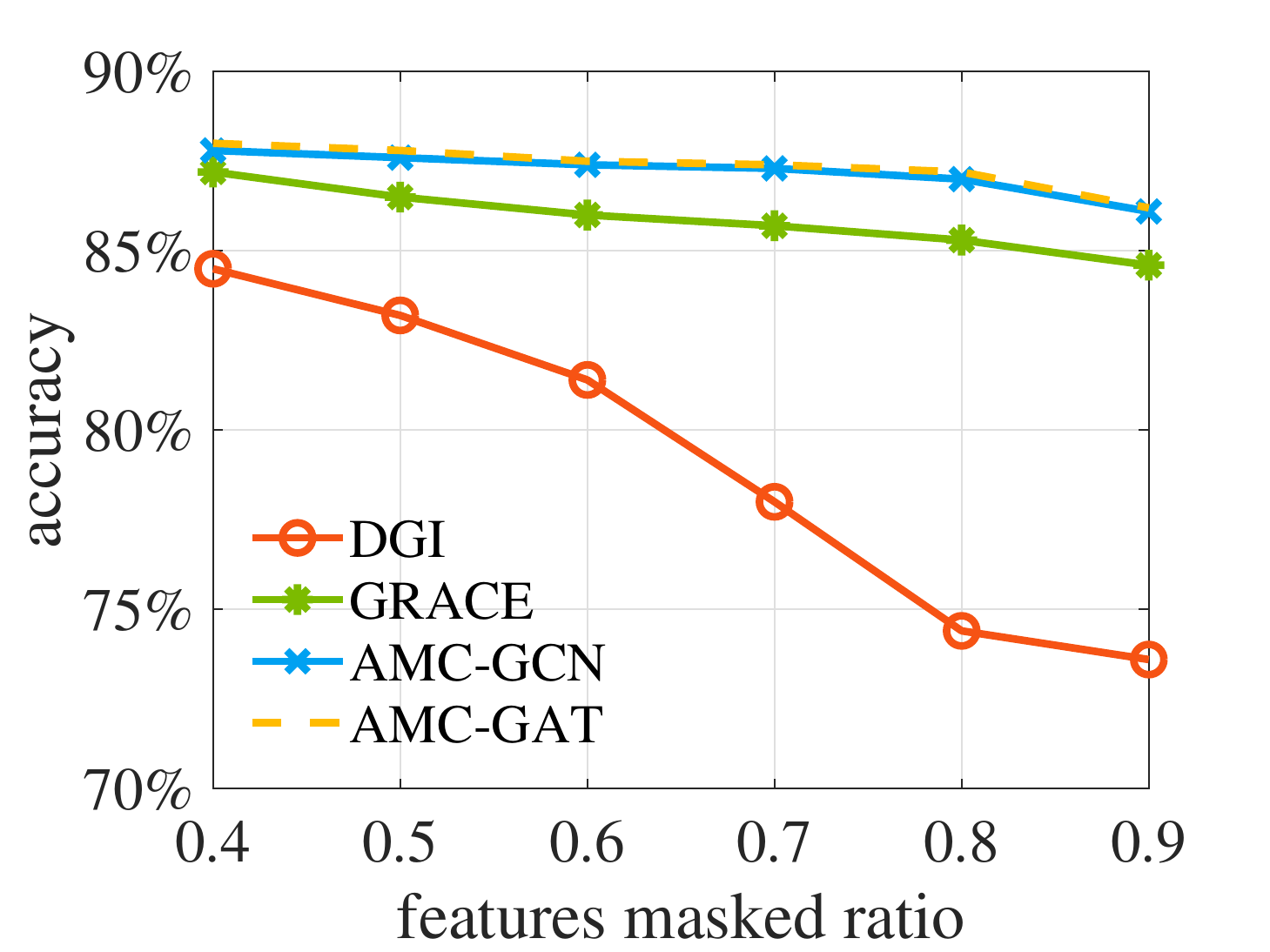}
\end{minipage}%
}
\caption{The performance of DGI, GRACE, AMC-GCN and AMC-GAT in transductive node classification on four datasets with different features masking ratio.}
\label{fig:6}
\centering
\end{figure}

The results on the four datasets are shown in Fig. \ref{fig:6}. With different pollution rates, AMC-GCN and AMC-GAT outperform GRACE and DGI in most cases, proving that our proposed AMC-GNN model has more robustness against the dropout of features. We attribute the robustness of AMC-GNN to the superiority of multi-layer comparison because the outputs of different layers are considered simultaneously, avoiding the cumulative propagation of errors caused by dropped features in the network. As the proportion of dropped features increases, the performance of the model decreases. This is because feature dropout has changed the semantic labels of the nodes, and the excessive dropout of node features prevents GNN from extracting meaningful information from the nodes.

\subsection{Representation Stability Visualization}
\label{sec:17}
\begin{figure}[h]
\centering
\subfloat[DGI]{
\begin{minipage}[h]{0.24\linewidth}
\centering
\includegraphics[width=1.4in]{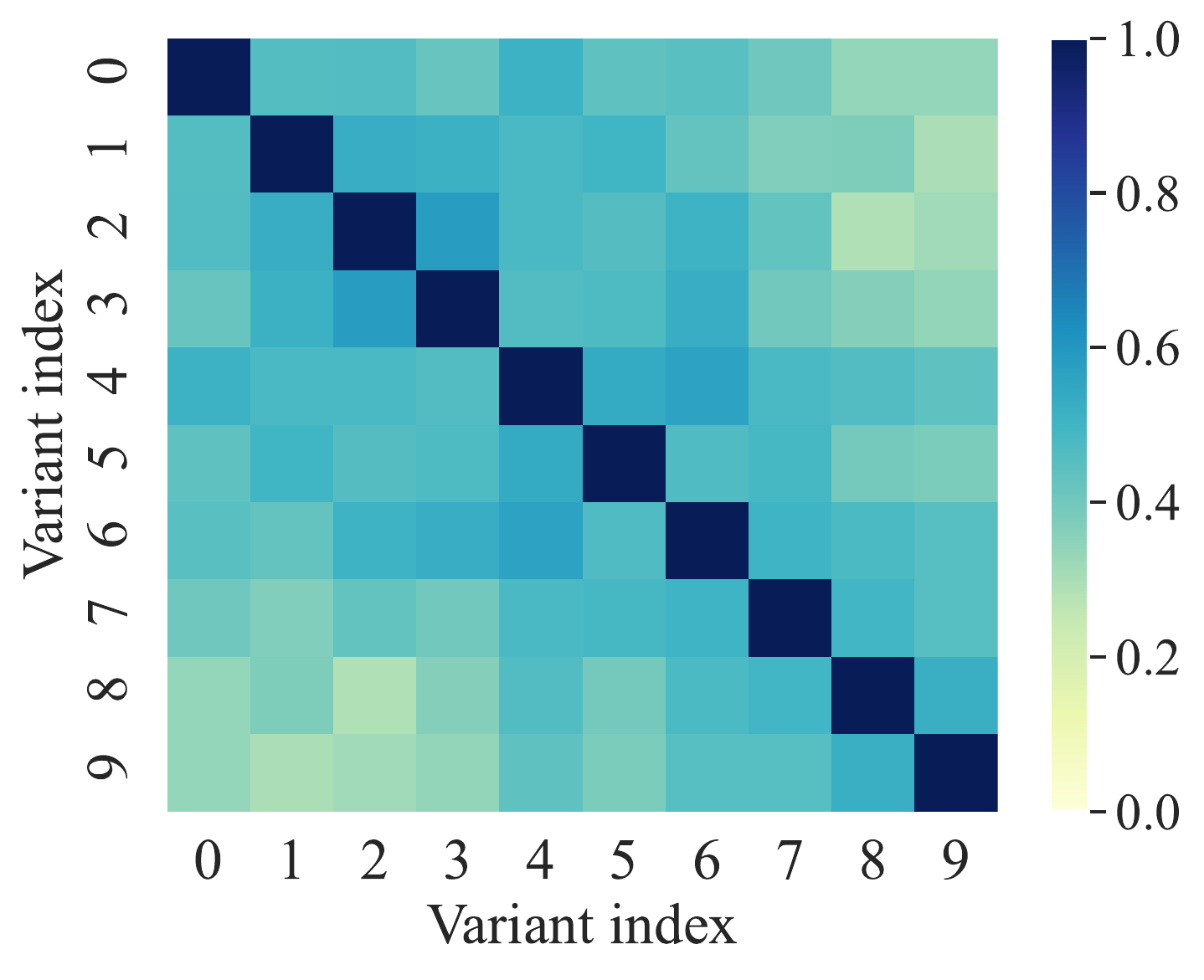}
\end{minipage}%
}%
\subfloat[GRACE]{
\begin{minipage}[h]{0.24\linewidth}
\centering
\includegraphics[width=1.4in]{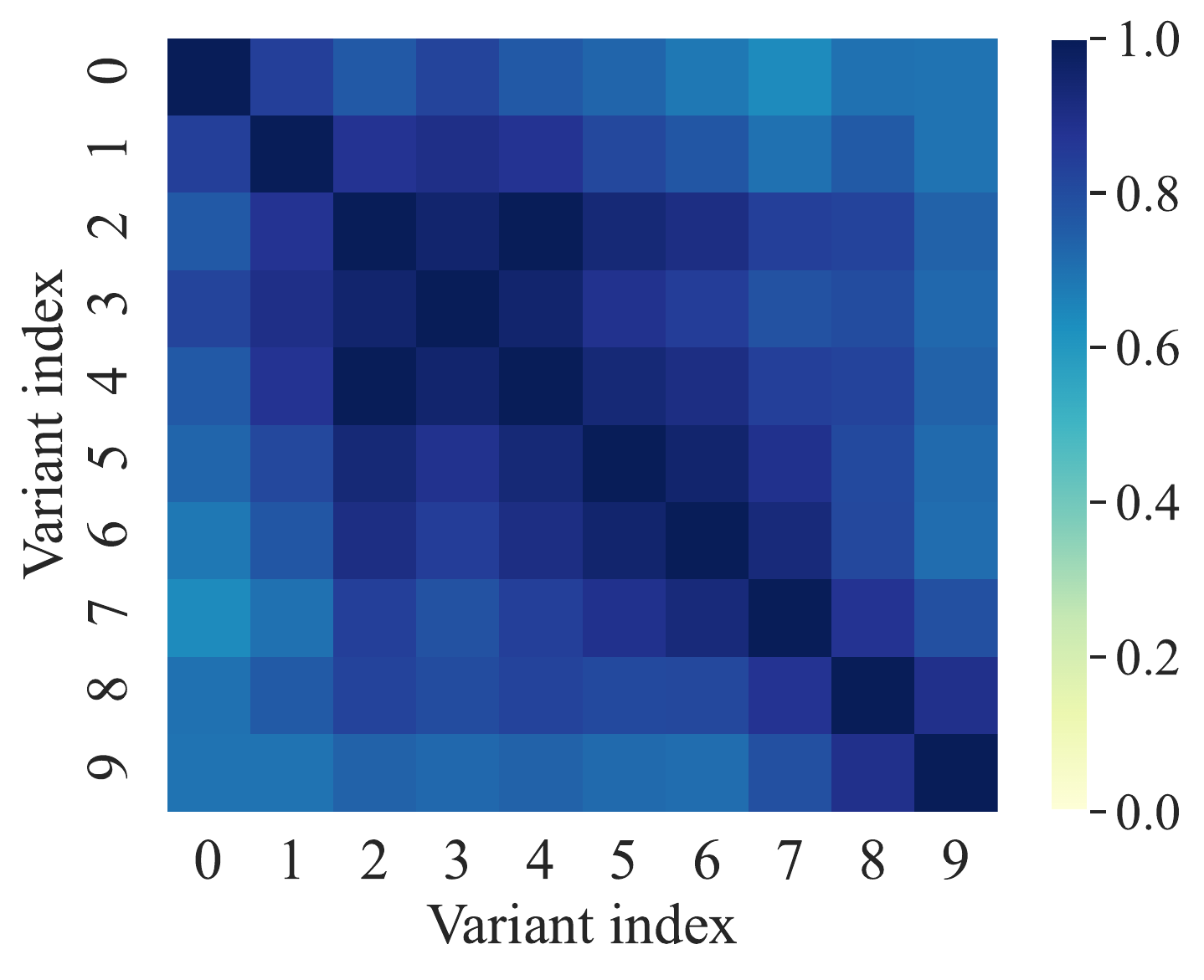}
\end{minipage}%
}%
\subfloat [AMC-GCN]{
\begin{minipage}[h]{0.24\linewidth}
\centering
\includegraphics[width=1.4in]{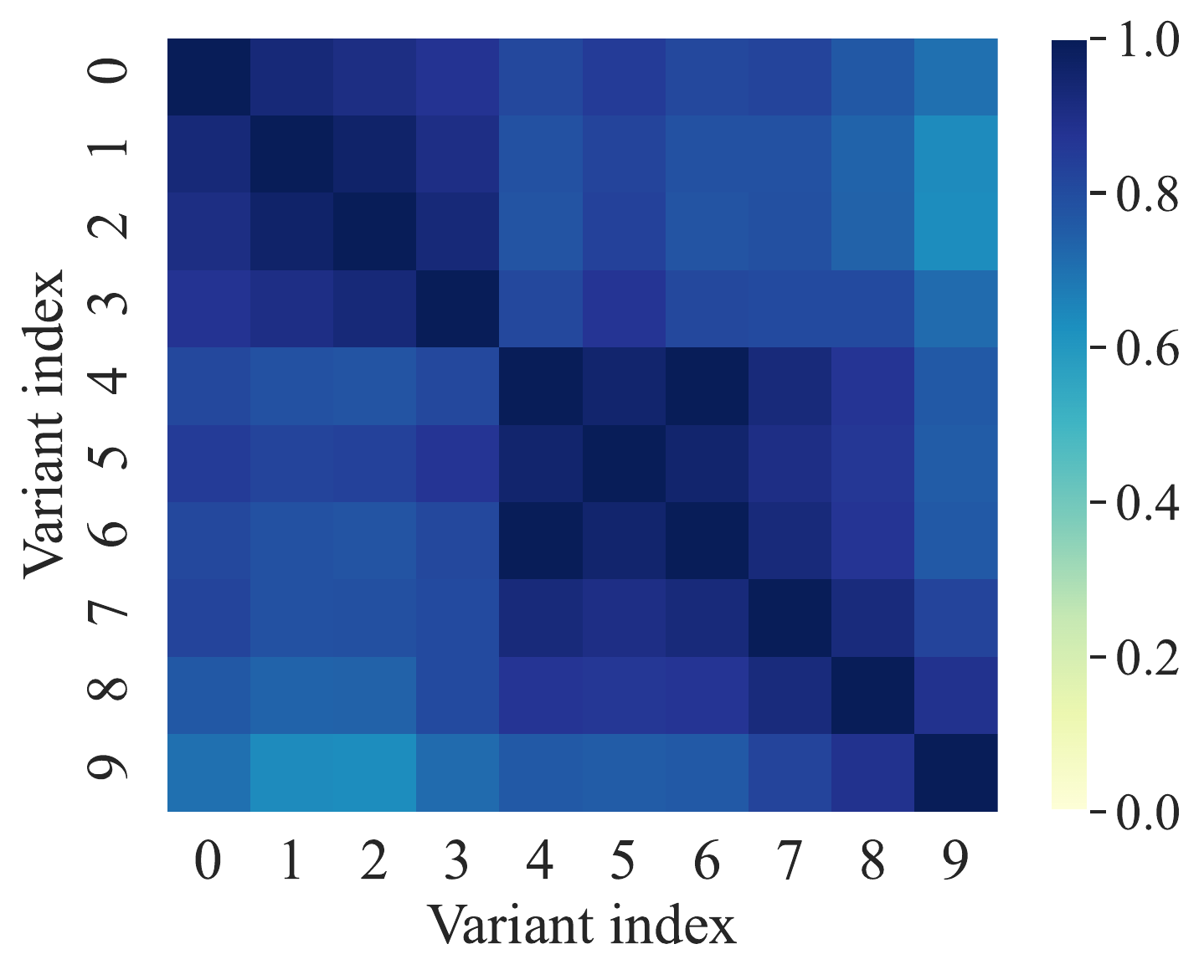}
\end{minipage}%
}%
\subfloat [AMC-GAT]{
\begin{minipage}[h]{0.24\linewidth}
\centering
\includegraphics[width=1.4in]{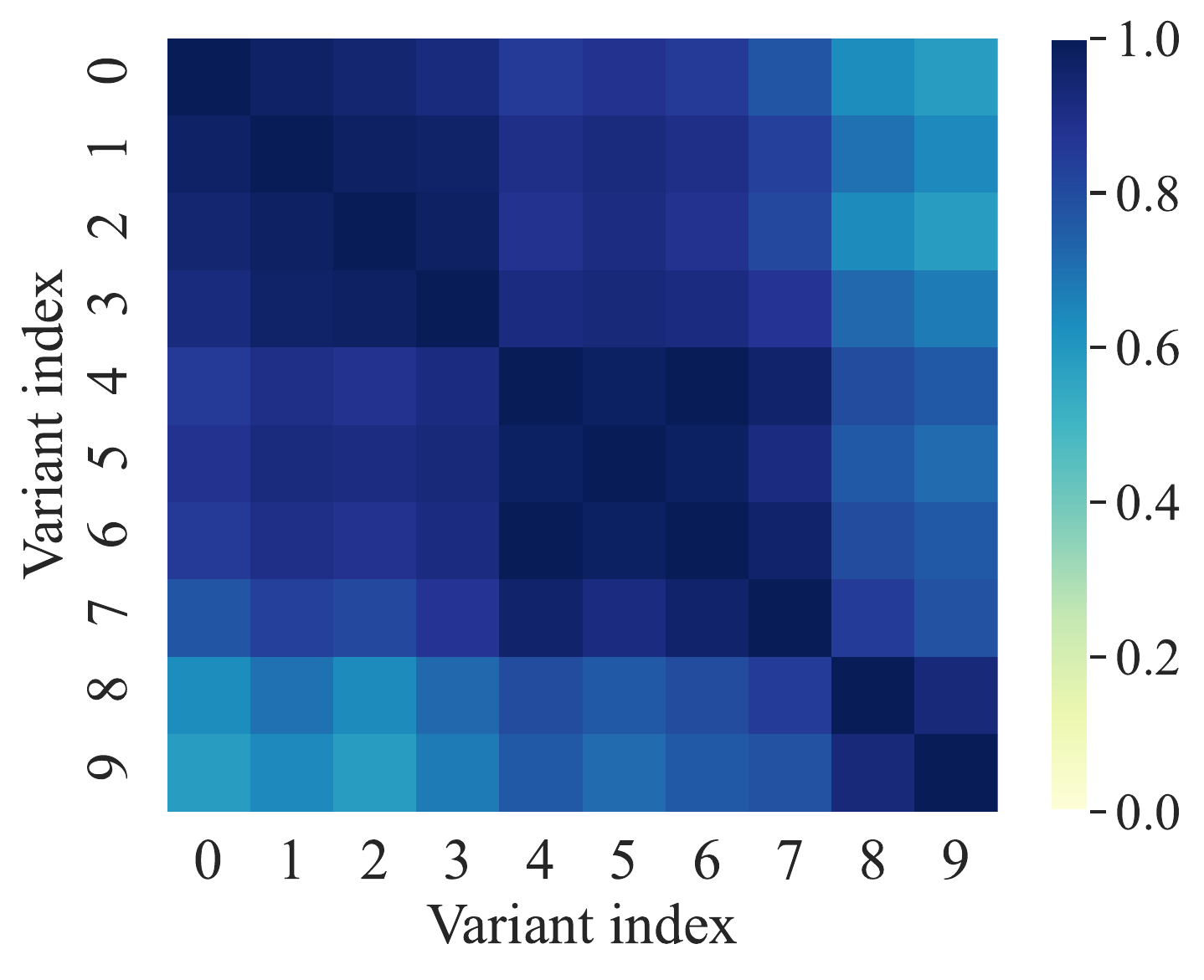}
\end{minipage}%
}%
\caption{The representation stability visualization on Cora. (a)(b)(c)(d) are performance of DGI, GRACE, AMC-GCN and AMC-GAT respectively. Deeper color indicates higher the similarity of the node representation.}
\label{fig:7}
\centering
\end{figure}

If the model is stable, then similar feature representations will be learned for nodes obtained by different data augmentation methods. We arbitrarily choose a node for 10 different dropping features with $p=0.4$ and get 10 different feature vectors ${\*{h}_{i}}$ after the trained target model, where $1\le i\le 10$.
Calculate the similarity matrix of the two sets of vectors $\*{S}\in {{\mathbb{R}}^{10\times 10}}$, where ${\*{S}_{ij}}=\frac{{\*{h}_{i}}\cdot {\*{h}_{j}}}{\left| {\*{h}_{i}} \right|\left| {\*{h}_{j}} \right|}$. The similarity matrix is visualized and the result are shown in Fig. \ref{fig:7}.

We randomly selected a test node to calculate the $\*{S}$ matrices for the four methods DGI, GRACE, AMC-GCN and AMC-GAT and visualize them in Fig. \ref{fig:7}. The similarity matrices of different models are visualized using darker colors to indicate higher similarity. Under the condition of slight disturbance, higher similarity of node features indicates better model stability and better extraction of essential features of nodes. The average similarity of DGI, GRACE, AMC-GCN, AMC-GAT are 0.503, 0.838, 0.849 and 0.859.

\section{Conclusion}
\label{conclusion}
In this paper, we propose a new contrastive GNN called AMC-GNN. AMC-GNN is a novel generic framework that could provide a new perspective on the structure of GNNs. AMC-GNN learns more essential features of different classes of data by introducing auxiliary training models and adding adaptive Multi-layer contrastive losses. AMC-GNN uses the attention mechanism to learn the importance weights of the embeddings in different layers adaptively. We conducted comprehensive experiments on various widely used datasets. The experimental results show that our proposed method can learn more robust and essential features of graphs, outperforming existing state-of-the-art unsupervised graph contrastive learning methods.


%
%

\bibliographystyle{spmpsci}      

\bibliography{MCGref}
\end{document}